\documentclass[10pt,journal,compsoc]{IEEEtran}
\usepackage{graphicx}
\usepackage{xcolor}
\usepackage{amsmath}
\usepackage{amssymb}
\usepackage{bm}
\usepackage{booktabs}
\usepackage{colortbl}
\usepackage{comment}
\usepackage{subfigure}
\usepackage{multirow}
\usepackage{hyperref}
\usepackage{color}
\ifCLASSOPTIONcompsoc
  \usepackage[nocompress]{cite}
\else
  \usepackage{cite}
\fi
\ifCLASSINFOpdf
\else
\fi
\usepackage{url}
\begin{document}
\title{
Pre-training Vision Transformers with Formula-driven Supervised Learning
}

%\author{Hirokatsu Kataoka$^{1}$, Ryo Hayamizu$^{1}$, Ryosuke Yamada$^{1}$, Kodai Nakashima$^{1}$, Sora Takashima$^{1,2}$,\\Xinyu Zhang$^{1,2}$, Edgar Josafat Martinez-Noriega$^{1,2}$, Nakamasa Inoue$^{1,2}$, Rio Yokota$^{1,2}$\\
%$^{1}$National Institute of Advanced Industrial Science and Technology (AIST)\\
%$^{2}$Tokyo Institute of Technology\\
%\footnotesize{\url{https://hirokatsukataoka16.github.io/Replacing-Labeled-Real-Image-Datasets/}}

\author{Hirokatsu~Kataoka,~Sora~Takashima,~Ryo~Hayamizu,~Ryosuke~Yamada,~Kodai~Nakashima,\\~Xinyu~Zhang,~Edgar~Josafat~Martinez-Noriega,~Nakamasa~Inoue,~Rio~Yokota% <-this % stops a space
\IEEEcompsocitemizethanks{
\IEEEcompsocthanksitem H. Kataoka, R. Hayamizu, R. Yamada, and K. Nakashima are with National Institute of Advanced Industrial Science and Technology (AIST).\protect\\
E-mail: hirokatsu.kataoka@aist.go.jp
\IEEEcompsocthanksitem S. Takashima, X. Zhang, E. J. Martinez-Noriega, N. Inoue, and R. Yokota are with National Institute of Advanced Industrial Science and Technology (AIST) and Institute of Science Tokyo.}% <-this % stops an unwanted space
\thanks{Manuscript received 25 December, 2025; revised December 25, 2025.}}


\markboth{Journal of \LaTeX\ Class Files,~Vol.~14, No.~8, August~2015}%
{Shell \MakeLowercase{\textit{et al.}}: Bare Demo of IEEEtran.cls for Computer Society Journals}



\IEEEtitleabstractindextext{%
\begin{abstract}
In the present work, we show that the performance of formula-driven supervised learning (FDSL) can match or even exceed that of ImageNet-21k and can approach that of the JFT-300M dataset \textbf{without} the use of real images, human supervision, or self-supervision during the pre-training of vision transformers (ViTs). For example, ViT-Base pre-trained on ImageNet-21k and JFT-300M showed 83.0 and 84.1\% top-1 accuracy when fine-tuned on ImageNet-1k, and FDSL showed 83.8\% top-1 accuracy when pre-trained under comparable conditions (hyperparameters and number of epochs). Especially, the ExFractalDB-21k pre-training was calculated with $\times14.2$ fewer images compared with JFT-300M. Images generated by formulas avoid privacy and copyright issues, labeling costs and errors, and biases that real images suffer from, and thus have tremendous potential for pre-training general models.
To understand the performance of the synthetic images, we tested two hypotheses, namely (i) object contours are what matter in FDSL datasets and (ii) an increased number of parameters for label creation improves performance in FDSL pre-training. To test the former hypothesis, we constructed a dataset that consisted of simple object contour combinations. We found that this dataset matched the performance of fractal databases. For the latter hypothesis, we found that increasing the difficulty of the pre-training task generally leads to better fine-tuning accuracy.
\end{abstract}

\begin{IEEEkeywords}
Pre-training, Vision Transformers, Formula-driven Supervised Learning, Fractal Geometry, Auto-generated Contours
\end{IEEEkeywords}}

\maketitle

\IEEEdisplaynontitleabstractindextext

% \fi
\IEEEpeerreviewmaketitle

\IEEEraisesectionheading{\section{Introduction}\label{sec:introduction}}

\IEEEPARstart{I}{mage} recognition has greatly benefited from labeled real-image datasets. Conventional image datasets comprise real images of various objects on a general background annotated by humans. A visual representation can be acquired by learning from real images with such annotations. Supervised learning (SL) is the most trusted approach for this task. These kinds of datasets with object labels usually rely on image collection and human annotation with significant human and time costs. Typically, the labeled real-image datasets have been constructed by a large number of crowd-workers categorized as collectors and annotators on the Internet. The crowdsourcing framework made it possible to realize the construction of large-scale real-image datasets in a relatively short period of time ever. Early in the deep learning era, the data scale was on the order of around 1M images. However, the dataset needed size to pre-train or train image recognition models is has reached over 100M images in the 2020s. Especially, a couple of huge-scale datasets, such as JFT-300M/3B~\cite{SunICCV2017_jft300m,ZhaiarXiv2021_ScalingViT} and IG-3.5B~\cite{MahajanECCV2018_ig3.5b}, have been validated in this research domain. The larger pre-training datasets have provided the most enhancement of performance rates of image recognition on downstream tasks. A pre-training dataset on the order of 100M labeled real images still improves the ability to classify image-level categories.

However, in recent years, self-supervised learning (SSL) has gained ground~\cite{HeCVPR2020,chen2020mocov2,ChenICML2020,chen2020big,ChenCVPR2021_simsiam}. The philosophy of SSL has become a promising approach in terms of unsupervised learning for visual representation without any object labels in the pre-training phase. Usually, SSL learning proceeds by generating consistent labels for input images. The study of SSL methods initially started with the objective of replacing SL on real images. As a result of the large number studies on SSL, the recognition performance has become much closer to the performance of SL with real-image datasets and manually annotated object labels. After a period when the SSL methods were proposed on convolutional neural networks (CNNs), these have recently been used to pre-train vision transformers (ViTs)~\cite{DosovitskiyICLR2021}; however, ViTs tend to require relatively large datasets in comparison to CNNs, such as datasets with tens or hundreds of millions of images~\cite{DengCVPR2009_ImageNet,SunICCV2017_jft300m}. 
% ViT / SSLの簡単な研究の略歴を紹介する
The SSL methods such as DeiT~\cite{TouvronICML2021}, DINO~\cite{CaronICCV2021_dino}, MoCoV3~\cite{ChenICCV2021_mocov3}, and masked auto-encoders (MAE)~\cite{HeCVPR2022_MAE} show that it is possible to train models on relatively small datasets such as ImageNet-1k (ILSVRC)~\cite{RussakovskyIJCV2015}.
SSL methods remove the time-consuming labeling of a dataset, but do not address the privacy, copyright, and societal biases that are present when real images are used~\cite{YangFAT2020,AsanoNeurIPS2021_pass}.

\label{sec:intro}
\begin{figure*}[t]
  \centering
  \includegraphics[width=0.90\linewidth]{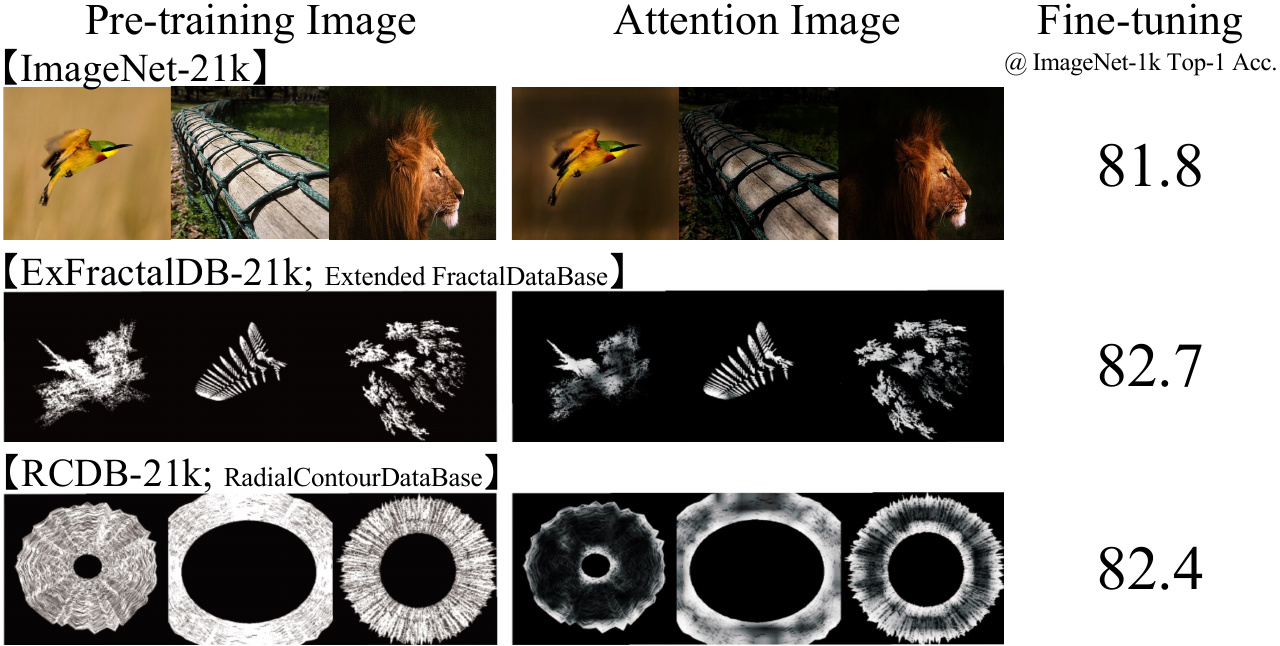}
  \caption{We have found that ViTs can be successfully pre-trained without real images, human supervision, or self-supervision, and can exceed the accuracy of ImageNet-21k pre-training when fine-tuned on ImageNet-1k. We constructed a new dataset, Radial Contour DataBase (RCDB), based on the assumption that contours are what matter for the pre-training of ViTs. Even though the ImageNet-21k pre-trained models have an overwhelming advantage in terms of ImageNet-1k fine-tuning because the pre-trained models have already looked at and trained on all images of the fine-tuning dataset, the Extended Fractal DataBase (ExFractalDB-21k) pre-trained model surpassed the ImageNet-21k pre-trained model. The ExFractalDB-21k pre-trained model has been improved from the original FractalDB pre-trained model based on two hypotheses presented in this paper. Moreover, RCDB also exceeded the performance of ImageNet-21k pre-training, while consisting only of contours.}
  \label{fig:figure1}
\end{figure*}

Formula-driven supervised learning (FDSL) trains on synthetic images generated by mathematical formulas and thus avoids such issues~\cite{KataokaACCV2020,NakashimaarXiv2021,InoueICPR2020,KataokaICCV2021WS,Baradad2021_deadleaves,AndersonWACV2022}.
The images can be categorized and labeled automatically based on the parameters of the equations used to generate them. 
The problem setting began with curiosity about the relationship between the training of visual representations and real-image datasets. Modern neural networks have allowed us to acquire sophisticated visual representations through large-scale image datasets with object labels. Here, the pixel space is undoubtedly enormous; for example, even a grayscale image with $224\times224$ ($50,176$) pixels makes $256^{50,176}$ dimensions in a feature space. This fact prompts us to employ a natural law that exists in the real world. Because images acquired by sensors are merely a cutout of the real world, the FDSL methods indicated that it would be better to use the laws behind the environment around us.
Because the images are generated by mathematical formulas, they avoid the ethical problems associated with labeled real-image datasets.
Although most of the current pre-training models have been created with real-images containing ethical issues, we do not have these concerns for the FDSL datasets. By using FDSL datasets, there is no chance for any ethical issues to arise, at least in the pre-training phase. The FDSL datasets do not have dataset biases or sensitive labels such as human race, gender, location.
For these reasons, if FDSL can be used to pre-train models to the same accuracy as real images, it could replace SL/SSL to avoid ethical issues.

To improve FDSL methods, Kataoka~\textit{et al.}~\cite{KataokaACCV2020} used fractal geometry based on the assumption that fractals are basic patterns found in nature. They found that the actual performance depends on the number of hyperparameters used to create FDSL datasets.
For example, is an image's fractal geometry the only important factor in a formula-driven framework for image pre-training? We recognize that this is one of the key issues in pre-training of image recognition networks.
In the present work, we investigated the most influential factors for generating synthetic images from formulas and the possibility of using alternatives to fractals. We established some basic guidelines that lead to better FDSL methods to avoid the iterative process of rendering and pre-training.

In this study, we enhanced the performance of FDSL in the context of pre-training ViTs~\cite{DosovitskiyICLR2021}. Throughout the large amount of experimental results and their consideration, we tested the following two hypotheses. Hypothesis 1: object contours are what matter in FDSL datasets. Hypothesis 2: an increased number of parameters effects performance improvement in FDSL pre-training. The preliminary study that led to these hypotheses is described in Section~\ref{sec:preliminary_study}. In the visualization of ViT self-attention related to hypothesis 1, models were highly responsive to the object contours in fractal images, rather than iteratively complicated fractal patterns. In the use of hyperparameters related to hypothesis 2, the fully randomized parameters performed better than the restricted parameters in FDSL pre-training. Through the validation of these hypotheses, we generated an improved synthetic dataset that allowed us to pre-train ViTs with an accuracy higher than that of real-image datasets. We summarize the main contributions in this paper as follows:

\noindent\textbf{\underline{Impact of this paper.}} We show that the performance of ViTs pre-trained with FDSL can match or even exceed that of ViTs pre-trained with ImageNet-21k. When fine-tuned on ImageNet-1k, ViT-Base pre-trained on ImageNet-21k had a top-1 accuracy of 81.8\% and ViT-Base pre-trained on the Extended Fractal DataBase (ExFractalDB) and Radial Contour DataBase (RCDB) (which has the same number of classes and instances per class) had accuracies of 82.7 and 82.4\%, respectively (Figure~\ref{fig:figure1}, Table~\ref{tab:comparison_imagenet1k}). It is surprising that not only the ExFractalDB-21k pre-trained model but also the RCDB pre-trained model mainly consisting of contour components outperformed ImageNet-21k which is currently a de-facto standard pre-trained dataset in the image recognition field. Moreover, to compare our proposed methods with the model pre-trained on JFT-300M, we additionally conducted an experiment on ViT pre-training. We changed the resolution setting in ImageNet-1k fine-tuning to 384$\times$384 pixels while the pre-training was unchanged at 224$\times$224 pixels. We confirmed that the ExFractalDB-21k and RCDB-21k pre-training with 384$\times$384 resolution respectively recorded 83.8 and 83.5\% in top-1 fine-tuning accuracy on the ImageNet-1k dataset. By considering that the JFT-300M pre-trained model produces an accuracy of 84.1\%, the scores with our FDSL pre-training are quite close to that of the pre-training on the largest real-image datasets.

\noindent\textbf{\underline{Effect of hypothesis 1.}} To understand the performance of fractal images, we explored alternative formulas for generating images. In our preliminary study (see Section~\ref{sec:preliminary_study}), we found that the object contours in the fractal images play an important role. Therefore, we created a dataset (RCDB) that is specifically tailored to drawing object contours. The performance of this dataset matches that of FractalDB (Table~\ref{tab:fdsl_labels}). Moreover, we revealed that the models pre-trained on ExFractalDB and RCDB with broken contours cannot successfully acquire feature representations in order to recognize objects on real-image datasets during the fine-tuning phase (Figure~\ref{fig:result_fractal_contour}), which validates our hypothesis.

\noindent\textbf{\underline{Effect of hypothesis 2.}} We find that higher complexity of the mathematically generated images improves the accuracy of FDSL (Table~\ref{tab:multiple_parameters}). The complexity of the images can be increased by adjusting the parameters of the formula-driven image generation. For example, RCDB becomes more complex when the number of vertices is increased; its complexity can also be changed by adjusting the contour smoothness, number of polygons, and radius (Table~\ref{tab:rcdb}). The complexity of FractalDB can be increased by applying an iterative function system (IFS) in three-dimensional (3D) space instead of two-dimensional (2D) space.

\section{Related work}
This paper is about pre-training ViTs for image recognition. Even if we limit the topic to image recognition and pre-training ViTs, it is closely related to several other important topics. Here, we mainly discuss image datasets, learning frameworks, and FDSL pre-training.
%Supervised training is the most reliable training mode in terms of accuracy. It is thus used as a baseline to measure the effectiveness of other training modes.
%Representative datasets such as ImageNet~\cite{DengCVPR2009_ImageNet,RussakovskyIJCV2015} and Places~\cite{ZhouTPAMI2017_Places} are collected, labeled, and cross-checked using cloud sourcing.

%SSL removes the annotation cost by automatically generating labels according to rules that can be learned ~\cite{DoerschICCV2015,ZhangECCV2016,NorooziECCV2016,NorooziCVPR2018,GidarisICLR2018}. The performance of contrastive SSL methods~\cite{HeCVPR2020,chen2020mocov2,ChenICML2020,chen2020big} is close to that of SL. For example, SimSiam~\cite{ChenCVPR2021_simsiam} can learn without negative samples and with smaller batch sizes. DINO~\cite{CaronICCV2021_dino} and MoCoV3~\cite{ChenICCV2021_mocov3} have demonstrated SSL on ViTs.

\subsection{Image datasets}

Over the last few decades, various image datasets have been proposed for training and testing computer vision models. Below we summarize datasets for image classification and object detection from the viewpoint of pre-training and fine-tuning neural networks.

For pre-training, a large-scale dataset is often required when a neural network has a large number of parameters. ImageNet, which consists of more than 14M images collected from the Internet, has been the most popular choice.
It provides ground-truth labels for 21k+ categories obtained using the Amazon Mechanical Turk. At the beginning of the deep learning era, the training set of ImageNet-1k, which contains 1.28M images in the ImageNet Large-Scale Visual Recognition Challenge (ILSVRC), has been frequently employed as the de-facto standard dataset for pre-training many visual tasks, such as image classification, object detection, and semantic segmentation.
Places~\cite{ZhouTPAMI2017_Places} is another large-scale dataset. It contains more than 10M images in 400+ scene categories.
Compared with ImageNet, Places is more specialized for scene comprehension; however, it is often effective for pre-training because many scene images contain various objects.
To avoid unintended biases, Pictures without humans for Self-Supervision (PASS) dataset provides a large number of images that do not include any humans~\cite{AsanoNeurIPS2021_pass}. To construct the privacy-preserving dataset, the detected faces and human bodies were automatically and manually eliminated from the pre-training dataset. Moreover, the dataset is assumed to be used for pre-training with self-supervision because the PASS dataset does not include manually assigned labels.
This dataset contains 1.4M images (a dataset scale similar to that of the ImageNet training, validation, and test datasets) available under a Creative Commons license (CC-BY).

Recently, we have witnessed a couple of massive-scale pre-training datasets which containing over 100M images for the computer vision field~\cite{BartACMCom2016,MahajanECCV2018_ig3.5b,SunICCV2017_jft300m,ZhaiarXiv2021_ScalingViT}. These massive-scale datasets have proved that a pre-trained visual model succeeds in transfer learning to a downstream task in the context of SL and SSL. Moreover, a 100M-scale dataset allows us to greatly improve visual models regardless of whether they are CNNs or ViTs. Collecting and labeling images at this scale is undoubtedly a challenging problem; however, a greater difficulty also lies in the fact that these datasets are limited to private usage. This severely limits the accessibility and reproducibility of research in computer vision.

Some small well-designed datasets are typically used for fine-tuning and evaluation.
For example, CIFAR-10 and CIFAR-100~\cite{Krizhevsky2009_cifar} are two of the most widely used datasets, each with 60k images for 10 and 100 categories, respectively.
The categories include animals and objects such as ``cat'', ``dog'', ``automobile'', and ``airplane''.
The Cars dataset consists of 16k images of 196 car categories~\cite{Krause3DRR2013_cars}.
The car categories are defined by the year and model such as ``2012 Tesla Model S'' and ``2012 BMW M3 coupe''.
The Flowers dataset consists of 102 flower categories~\cite{Nilsback08_flowers}.
The Pascal Visual Object Classes (VOC) dataset is used as an image classification benchmark in addition to object detection and semantic segmentation~\cite{EveringhamIJCV2015_voc}.
The VOC 2007 and VOC 2012 datasets contain 9,963 and 11,530 images, respectively, in 20 classes.
ImageNet-100 and Places-30 are also employed to evaluate the recognition accuracy with a pre-trained model. These are subsets of the ImageNet-1k and Places-365 datasets as described at the beginning of this subsection. These datasets are roughly one-tenth the size of the original ImageNet-1k and Places-365.
Fine-tuning can also be applied to object detection, aiming to output bounding boxes for each object in an image. Large-scale object detection datasets include COCO~\cite{LinECCV2014_coco} and Open Images~\cite{openimages}.
It is worth noting that these datasets have official subsets for training and testing. This enables researchers to fairly compare the performance of different methods.

\subsection{Learning frameworks}
% SL, SSL, FDSL
Supervised learning is the most reliable training mode in terms of accuracy. Starting with the ImageNet pre-trained AlexNet, we have observed several representative models that were trained with SL~\cite{alexnet,vggnet,inception,resnet,resnext,densenet,senet,efficientnet}. This is thus used as a baseline to measure the effectiveness of other training modes.
However, SL requires a large effort and cost because manually attached labels are needed for all training images.

Self-supervised learning has recently proven to be effective for learning image representations from unlabeled image datasets.
The basic idea of SSL is to train neural networks on a pre-text task such as context encoder~\cite{DoerschICCV2015}, jigsaw puzzle~\cite{NorooziECCV2016,NorooziCVPR2018}, image rotation~\cite{GidarisICLR2018}, and colorization~\cite{ZhangECCV2016}.
Given a pre-text task, a neural network learns to solve the task while learning image representations in an unsupervised manner.
The performance of SSL using such pre-text tasks is typically worse than that of SL, but the idea of using unlabeled images has proven to be useful.

The most recent trend in SSL is to use a contrastive loss defined over augmented images and MAEs by creating deficiencies in image patches.
For example, SimCLR~\cite{ChenICML2020} introduces NT-Xent loss based on noise contrastive estimation. It makes pairs of augmented images in a mini-batch and predisposes a neural network to solve the problem of whether paired images are obtained from the same image.
MoCo~\cite{HeCVPR2020} and the improved MoCov2/v3~\cite{chen2020mocov2, ChenICCV2021_mocov3} use a momentum encoder obtained by taking the moving average of an encoder for improving contrastive loss.
Their variants further improve the performance of SSL.
Examples include BYOL~\cite{GrillNeurIPS2020_byol}, Barlow Twins~\cite{ZbontarICML2021_BT}, SimSiam~\cite{ChenCVPR2021_simsiam} and DINO \cite{CaronICCV2021_dino}.
However, a counterpart of contrastive SSL is proposed for MAEs~\cite{HeCVPR2022_MAE}. Based on BERT~\cite{Devlin2018bert} with reference to the field of natural language processing, an MAE acquires visual representations while making masked patches in an image and reconstructs them like auto-encoders. 
It is worth noting that MAEs are comparable with or even better than SL on some computer vision tasks, including image classification and object detection.
For the recent ViTs, an MAE~\cite{HeCVPR2022_MAE} is one of the most promising methods.

\subsection{FDSL Pre-training}
Popular datasets have privacy and fairness issues, such as the ethical problems~\cite{YangFAT2020,PrabhuWACV2021_imagedatasets} in ImageNet (human-related labels)~\cite{DengCVPR2009_ImageNet} and 80M Tiny Images~\cite{TorralbaTPAMI2008}, which have led to the suspension of their publication~\footnote{\url{https://groups.csail.mit.edu/vision/TinyImages/}}.
SSL can eliminate labeling cost but it does not address ethical issues.
Even PASS~\cite{AsanoNeurIPS2021_pass}, a database of images available under a CC-BY license that excludes images of people, might have some harmful content.
It is a well-established fact that basic visual abilities can be acquired by learning a dataset composed of real images of various objects captured in various scenes, to which humans are assigned paired training labels.
FDSL is based on the hypothesis that the use of real images and manual annotations is actually the cause of artificial intelligence ethics harm.
Image patterns are generated from a mathematical formula based on natural principles, without using any real images or manual annotations.
FDSL is very challenging due to the lack of both manual image collection and manual annotation.
However, there are a number of FDSL methods~\cite{KataokaACCV2020,InoueICPR2020,KataokaICCV2021WS,YamadaIROS2021,NakashimaarXiv2021,Baradad2021_deadleaves,AndersonWACV2022,BaradadNeurIPS2022_shader,takashima2023visual,nakamura2023pretraining,nakamura2024scaling,hayamizu2024sifter,yamada2024formula,matsuo2025moiredb} showing the possibility of training with synthesized images.
In this study, we confirmed that our proposed FDSL datasets reached the accuracy of a de-facto standard real image dataset including ImageNet-21k pre-training.
Nakashima \textit{et al.} succeeded in pre-training ViTs on FractalDB by matching the accuracy of SSL (SimCLRV2)~\cite{NakashimaarXiv2021}.
However, they did not analyze the essence of FDSL, investigate its failure modes, or scale up to the size of ImageNet-21k.
In the present work, we performed a wider search of possible FDSL methods as well as larger-scale FDSL datasets, and analyzed the characteristics that correlate with good pre-training performance.
We also explored the range of parameters and configurations for which FDSL completely fails, providing insight into what constitutes a favorable synthetic image dataset for pre-training ViTs.

\section{Method}
We first show the results of our preliminary study from which we deduced the hypotheses that (i) object contours are what matter in image representation and (ii) increasing the number of parameters improves the performance of FDSL pre-training.
Then, we describe a set of mathematically generated datasets that have varying degrees of complexity in these two aspects to verify our hypotheses.

To verify the hypothesis regarding the importance of object contours in images, we created the dataset RCDB. To verify the second hypothesis, we increased the number of parameters in the equations used to generate the images in ExFractalDB and RCDB and increased the size of the dataset.

\begin{figure}[t]
  \centering
  \includegraphics[width=1.0\linewidth]{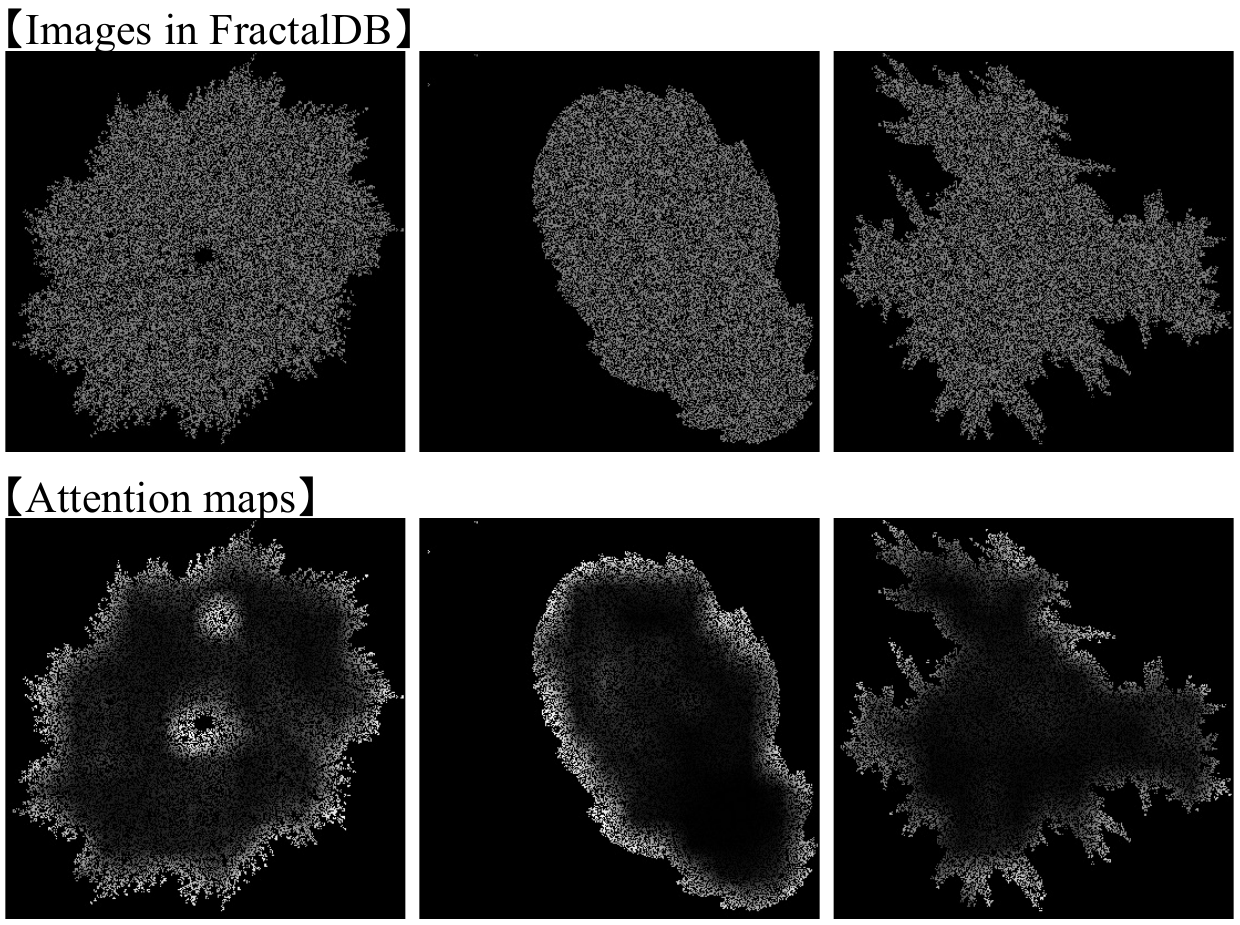}
  \caption{Fractal images and attention maps on object contours. The first and second columns respectively show original images on FractalDB and attention maps in self-attention on the model. According to these figures, ViTs apparently focus on outer contours in fractal images while acquiring visual representation in pre-training phase. We developed hypothesis 1 in this paper with a preliminary study.}
  \label{fig:contour_selfattention}
\end{figure}

\subsection{Preliminary study}
\label{sec:preliminary_study}
As a baseline, we first report the results for FractalDB-1k~\cite{KataokaACCV2020}, which has 1k classes and 1k instances per class.
We selected ViT-Tiny with $16\times16$ pixel patches as the baseline model. The hyperparameters and data augmentation were selected according to previous work~\cite{TouvronICML2021}.

\noindent \textbf{Hypothesis 1: Object contours are what matter in FDSL datasets}. The attention map for the training of ViTs with FractalDB-1k is shown in Figure~\ref{fig:contour_selfattention}, where attention is focused on the outer contours of the fractals.
The same behavior was observed across other images in the dataset.
We previously believed that the ability of fractals to generate recurring patterns found in nature is what allows them to be used as alternatives to real images.
However, our preliminary experiments suggested that the same effectiveness can be achieved by generating object contours with a sufficiently high complexity.\\

\begin{table}[t]
    \begin{center}
    \caption{Relationship between FractalDB-1k and labeling. FDSL (Fractal) corresponds to the original FractalDB, and FDSL (Fractal, restricted) varies only three parameters $(a_{i}, c_{i}, e_{i})$ while holding the other three $(b_{i}, d_{i}, f_{i})$ fixed. Best values are in bold.}
    \begin{tabular}{llcccc} \toprule[0.8pt]
        Type & C10 & C100 & Cars & Flowers \\\midrule[0.5pt]
        SSL \scriptsize{(MoCov2)} & 92.6 & 73.7 & 33.6 & 93.9 \\ 
        SSL \scriptsize{(SimCLRv2)} & 94.8 & 78.9 & 61.7 & \textbf{99.6} \\
        FDSL \scriptsize{(Fractal, restricted)} & 96.8 & 82.0 & 86.8 & 98.2 \\
        \rowcolor[gray]{0.8} FDSL \scriptsize{(Fractal)} & \textbf{97.0} & \textbf{82.4} & \textbf{87.9} & 98.3 \\\bottomrule[0.8pt]
    \end{tabular}
    \label{tab:fdb1k_labels}
    \end{center}
\end{table}

\begin{table}[t]
    \begin{center}
    \caption{Parameters in restricted FractalDB. The three parameters $(b_{i}, d_{i}, f_{i})$ are fixed to generate a fractal image.}
    \begin{tabular}{cccccc} \toprule[0.8pt]
        a & b & c & c & e & f \\\midrule[0.5pt]
        0.33	& 0.44	& 0.63	& -0.39	& -0.7	& -0.81 \\
        0.011	& 0.44	& -0.91	& -0.39	& 0.97	& -0.81 \\
        -0.007	& 0.44	& -0.64	& -0.39	& 0.72	& -0.81 \\
        0.29	& 0.44	& -0.13	& -0.39	& 0.097	& -0.81 \\
        0.23	& 0.44	& -0.81	& -0.39	& -0.32	& -0.81 \\
        -0.2	& 0.44	& 0.019	& -0.39	& -0.23	& -0.81 \\
        0.4	& 0.44	& -0.44	& -0.39	& -0.16	& -0.81 \\
    \bottomrule[0.8pt]
    \end{tabular}
    \label{tab:fractal_restricted}
    \end{center}
\end{table}

%\subsubsection{Role of supervision in FDSL}
%\noindent \textbf{Hypothesis 2: FDSL parameters can be adjusted to make pre-training difficult.}
\noindent \textbf{Hypothesis 2: Increasing the number of parameters in FDSL pre-training}. For both FDSL and SSL, manual labeling is unnecessary.
FDSL automatically labels the dataset during its creation, which is fundamentally different from SSL.
To investigate the role of labels in FDSL, we compared the pre-training on fractal images with FDSL and SSL, respectively (Table~\ref{tab:fdb1k_labels}).
For SSL, we selected MoCoV2 and SimCLRV2.
We compared two types of FractalDB, the original method with six parameters, FDSL (Fractal), and another method that varies only three parameters while holding the other three fixed, FDSL (Fractal, restricted). Table~\ref{tab:fractal_restricted} shows the correspondences, where a subset of the parameters in $\theta$ of IFS are fixed. The FDSL (Fractal, restricted) varies only three parameters $(a_{i}, c_{i}, e_{i})$ and holds the other three $(b_{i}, d_{i}, f_{i})$ fixed. Table~\ref{tab:fractal_restricted} shows an example of the restricted parameters. The three parameters $(a_{i}, c_{i}, e_{i})$ that we changed were set to match those in the FractalDB paper~\cite{KataokaACCV2020}.

Our results show that FDSL yields higher accuracy than does SSL when using FractalDB.
This result indicates that it is better to use labels from a mathematical formula that generates an image pattern with FDSL, than to assign an external label with SSL.
We also find that FractalDB created with a larger number of parameters leads to higher accuracy.
We therefore hypothesized that the accuracy of FDSL can be improved by increasing the parameters in equations to create FDSL labels.

\subsection{Formula-driven supervised learning}
\begin{figure*}[t]
  \centering
  \includegraphics[width=0.95\linewidth]{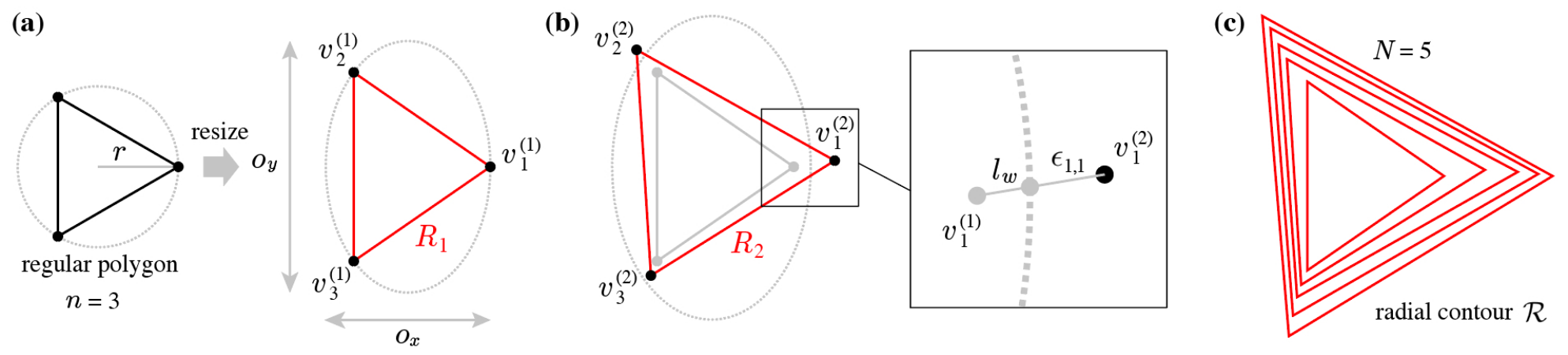}
  \caption{Procedure for generating radial contours $\mathcal{R}$. An example with $n=3$ vertices and $N=5$ polygons is shown.}
  \label{fig:radial}
\end{figure*}

FDSL automatically generates image patterns and their corresponding labels based on formulas.
Unlike the pre-training in SL/SSL, FDSL does not require real images.

\noindent \textbf{Definition of FDSL.}
Let $\phi_{\theta}$ be a network to be pre-trained with a parameter set $\theta$. FDSL solves the following problem:
\begin{align}
  \label{eq:fdsl}
  \hat \theta = 
  \mathop{\rm argmin}_{\theta}\limits
  \mathbb{E}_{({\bm x}, y)} [ \mathcal{L}(\phi_{\theta}({\bm x}), y) ]
\end{align}
where ${\bm x}$ is the synthetic image, $y$ is the corresponding label, and $\mathcal{L}$ is a loss function.
The synthetic images are generated by ${\bm x} = F_{y}(s)$,
where $y \in \{1,2,\cdots,C\}$ is a discrete label and $F_{y}$ is the $y$-th mathematical formula used to generate intra-class images.
Note that to create intra-class variation, $F_{y}$ involves randomness, so a random seed $s$ is input.
This leads to a two-step sampling of $({\bm x}, y)$ in Eq.~(\ref{eq:fdsl}), where $y$ is first uniformly sampled and then image pattern ${\bm x}$ is generated with $F_{y}(s)$ using the uniformly sampled seed $s$.

\subsubsection{Radial Contour Database (RCDB)}

\noindent \textbf{Definition of RCDB.}
The proposed radial contour $\mathcal{R} \subset \mathbb{R}^{2}$ is an object made by superimposing polygons. It is defined by a union set of polygons as follows:
\begin{align}
 \mathcal{R} = \bigcup_{p=1}^{N} R_{p},
\end{align}
where $R_{p}$ is the $p$-th polygon and $N$ is the number of polygons.
Each polygon $R_{p}$ consists of $n$ edges as follows:
\begin{align}
 R_{p} = \bigcup_{j=1}^{n} e({\bm v}^{(p)}_{j-1}, {\bm v}^{(p)}_{j})
\end{align}
where ${\bm v}^{(p)}_{j} \in \mathbb{R}^{2}$ is the $j$-th vertex. Note that we define vertices for $j = 0, 1, \cdots, n$, but ${\bm v}^{(p)}_{0} = {\bm v}^{(p)}_{n}$ is redundant. $e(\cdot, \cdot)$ is the edge between two vertices given by
\begin{align}
\label{eq:edge}
 e({\bm p}, {\bm q}) =
 \{ t {\bm p} + (1 - t) {\bm q} + {\bm c} \in \mathbb{R}^{2}: 0 \leq t \leq 1\}
\end{align}
where ${\bm c}$ is the center of the polygon.

Our algorithm makes polygons from the center to the border as follows.
The first polygon $R_{1}$ is made by resizing an $n$-regular polygon, that is, the vertices are given by
\begin{align}
{\bm v}^{(1)}_{j} = r
\begin{pmatrix}
o_{x} \cos \left( 2 \pi j / n \right) \\
o_{y} \sin \left( 2 \pi j / n \right) \\
\end{pmatrix}
\end{align}
for $j = 0, 1, 2, \cdots, n$ with radius $r$ and a resizing factor ${\bm o} = (o_{x}, o_{y})$. This step is illustrated in Figure~\ref{fig:radial}(a).

In the second step, vertices are copied and moved toward the border.
Specifically, given the vertices for $R_{p-1}$, new vertices for $R_{p}$ are defined as
\begin{align}
{\bm v}^{(p)}_{j} = {\bm v}^{(p-1)}_{j} + 
\begin{pmatrix}
(l_{w} + \lambda_{x} \epsilon_{j,p-1}) \cos \left( 2 \pi j / n \right) \\
(l_{w} + \lambda_{y} \epsilon_{j,p-1}) \sin \left( 2 \pi j / n \right) \\
\end{pmatrix},
\end{align}
where $l_{w}$ is the line width,
${\bm \epsilon}_{j} = (\epsilon_{j,1}, \epsilon_{j,2}, \cdots, \epsilon_{j,N})$ is a one-dimensional Perlin noise sequence,
and ${\bm \lambda} = (\lambda_{x}, \lambda_{y})$ is a noise scaling factor. This step is repeated for $p = 2 , 3, \cdots, N$.
Examples of $R_{2}$ and $\mathcal{R}$ are shown in Figures~\ref{fig:radial}(b) and (c), respectively.
Finally, the radial contour is rendered with a white line with a width of $l_{w}$ over a black background. The image size is $512\times512$ pixels.

\begin{table}[t]
    \begin{center}
    \caption{Parameter set ($\eta$) for RCDB categories.}
    \begin{tabular}{ll} \toprule[0.8pt]
        Parameter set ($\eta$) &  Values \\
        \midrule[0.1pt]
        \# of polygons ($N$) & \{1, 2, 3, ... , 200\} \\
        \# of vertices ($n$) & \{3, 4, 5, ... , 502\} \\
        Radius ($r$) & [0.0, 100.0] \\
        Line width ($l_{w}$) & [0.0, 0.1] \\
        Resizing factor (${\bm o}$) & [1.0, 4.0] \\
        Perlin noise (${\bm \lambda}$) & [0.0, 4.0] \\
        \bottomrule[0.8pt]
    \end{tabular}
    \label{tab:rcdb}
    \end{center}
\end{table}

\noindent \textbf{RCDB-1k.}
Let $\eta = (N, n, r, l_{w}, {\bm o}, {\bm \lambda})$ be a hyperparameter set for generating radial contours.
The proposed database consists of $C =$ 1k radial contour classes, each with a parameter set $\eta_{y} \ (y \in \{1,2,\cdots, C\})$. With this generation procedure denoted as $G_{\text{RC}}$, the definition of $F_{y}$ in this database is given by $F_{y}(s) = G_{\text{RC}}(\eta_{y}, s)$,
where random seed $s$ is used to randomly choose the center ${\bm c}$ in Eq.~(\ref{eq:edge}) and generate one-dimensional Perlin noise sequences. The hyperparameters are uniformly distributed over the range shown in Table~\ref{tab:rcdb}. For each class, 1k images are generated.

\noindent \textbf{Scaling RCDB.}
To explore the possibility of large-scale RCDB pre-training,
we prepared three more databases with the number of classes $C=$ 10k, 21k, and 50k.
The number of images per class was set to 1k for all databases.

\subsubsection{Extended FractalDB (ExFractalDB)}
\noindent \textbf{FractalDB.}
The original FractalDB proposed in \cite{KataokaACCV2020} consists of 2D fractal images generated by an IFS~\cite{fractals_everywhere}.
It has $C$ fractal classes, each of which has a hyperparameter set $\eta_{y}$ to generate fractals as $F_{y}(s) = G_{\text{IFS}}(\eta_{y}, s)$,
where $G_{\text{IFS}}$ is the rendering procedure based on IFS and $s$ is a random seed used to create intra-class variation.

The IFS is defined by
\begin{align}
\text{IFS}
=
\{
\mathcal{X};
w_{1}, w_{2}, \cdots, w_{N};
p_{1}, p_{2}, \cdots, p_{N};
\},
\end{align}
where $\mathcal{X}$ is a complete metric space,
$w_{i}: \mathcal{X} \to \mathcal{X}$ are transformation functions,
$p_{i}$ is a probability mass function,
and $N$ is the number of transformations.
When the transformation functions are parameterized,
the hyperparameter set $\eta_{y}$ is given by
\begin{align}
\eta_{y} = \{(\alpha_{i}, p_{i})\}_{i=1}^{N},
\end{align}
where $\alpha_{i}$ is the parameter set for the transformation function $w_{i}$. It is worth noting that the hyperparameter set $\eta_{y}$ is pre-defined  for each fractal class $y$, and this characterizes the fractal shapes.

The rendering procedure $G_{\text{IFS}}$ generates fractal images by using the random iteration algorithm~\cite{fractals_everywhere}.
Specifically, given the IFS with a hyperparameter set $\eta_{y} = \{(\alpha_{i}, p_{i})\}_{i=1}^{N}$,
a fractal $\mathcal{S} = \{\bm{v}_{t}\}_{t=1}^{\infty} \subset \mathcal{X}$ is constructed by repeating the following two steps for $t = 0, 1, 2, \cdots, T$.
(1) Select a transformation function $w^{*}$ from $\{w_{i}\}_{i=1}^{N}$ with probabilities $p_{i} = p(w^{*} = w_{i})$. (2) Produce a new point $\bm{v}_{t+1} = w^{*}(\bm{v}_{t})$.
Note that the initial point $\bm{v}_{0}$ is the center of an image. Each point is drawn in white on a black background.

Each transformation function is assumed to be an affine transformation in the 2D Euclidean space $\mathcal{X} = \mathbb{R}^{2}$ given by
\begin{align}
\label{eq:affine2d}
w(\bm{v}; \alpha)
=
\begin{pmatrix}
A_{11} & A_{12}\\
A_{21} & A_{22}
\end{pmatrix}
\bm{v}
+
\begin{pmatrix}
b_{1}\\
b_{2}
\end{pmatrix},
\end{align}
where $\bm{v} \in \mathbb{R}^{2}$ is a point and
$\alpha = (A_{11}, A_{12}, A_{21}, A_{22}, b_{1}, b_{2})$ is the parameter vector for 2D affine transformation.  

The FractalDB consists of $C=$1k classes. For each class, the number of transformations $N$ is sampled from the discrete uniform distribution on $\{2,3,4,5,6,7,8\}$.
Each element of the parameter vector $\alpha_{i}$ is sampled from the uniform distribution on $[-1,-1]$.
The probability $p_{i}$ is set to $\text{det}\bm{A}^{(i)}/\sum_{i}(\text{det} \bm{A}^{(i)})$, where $\bm{A}^{(i)} = (A^{(i)}_{11}, A^{(i)}_{12}; A^{(i)}_{21}, A^{(i)}_{22})$.

\noindent \textbf{ExFractalDB.}
MV-FractalDB~\cite{YamadaIROS2021} consists of 2D images that are projections of 3D fractals.
The fractals are generated by 3D-IFS, which replaces the 2D affine transformation function in Eq.~\ref{eq:affine2d} with a 3D function:
\begin{align}
\label{eq:affine3d}
w(\bm{v}; \alpha)
=
\begin{pmatrix}
A_{11} & A_{12} & A_{13}\\
A_{21} & A_{22} & A_{23}\\
A_{31} & A_{32} & A_{33}
\end{pmatrix}
\bm{v}
+
\begin{pmatrix}
b_{1}\\
b_{2}\\
b_{3}
\end{pmatrix},
\end{align}
where $\bm{v} \in \mathbb{R}^{3}$.
The generated 3D fractals are projected onto 2D images via a virtual camera.

\begin{table*}[h]
    \begin{center}
    \caption{Definitions of class and instance in Fractals (FractalDB, MV-FractalDB, and ExFractalDB) and Radial Contours (\#Vertices and parameter set $\eta$).}
    \begin{tabular}{lll} \toprule[0.8pt]
        Database & Class & Instance \\\midrule[0.5pt]
        FractalDB~\cite{KataokaACCV2020} & Random parameters from 2D-IFS & Image rotation, $3\times3$ patch pattern, weighting values \\
        MV-FractalDB~\cite{YamadaIROS2021} & Random parameters from 3D-IFS & Fixed viewpoints \\
        ExFractalDB* & Random parameters from 3D-IFS & Random viewpoints \\\midrule[0.5pt]
        RCDB & \#Vertices & \#Polygon, radius, line width, resizing factor, Perlin noise \\
        RCDB* & Parameter set $\eta$ & Image translation \\\midrule[0.5pt]
        Dead Leaves~\cite{Baradad2021_deadleaves} & Ratios of circle, triangle, and square & Randomly located circles, triangles, and squares \\ \midrule[0.5pt]
        Bezier Curves~\cite{KataokaACCV2020} & \#Lines \& \#Dots & Randomly located lines and dots \\
        Bezier Curves* & \#Control points,  radius, resizing factor &  \#Lines, vectors with connecting control points \\ \midrule[0.5pt]
        LineDB & \#Lines & Randomly located lines  \\
        \bottomrule[0.8pt]
        \multicolumn{3}{l}{* The FDSL datasets were enhanced based on hypothesis 2 in this paper.}\\
    \end{tabular}
    \label{tab:fdsl_class_instance}
    \end{center}
\end{table*}

\begin{figure*}[h]
  \centering
  \includegraphics[width=0.9\linewidth]{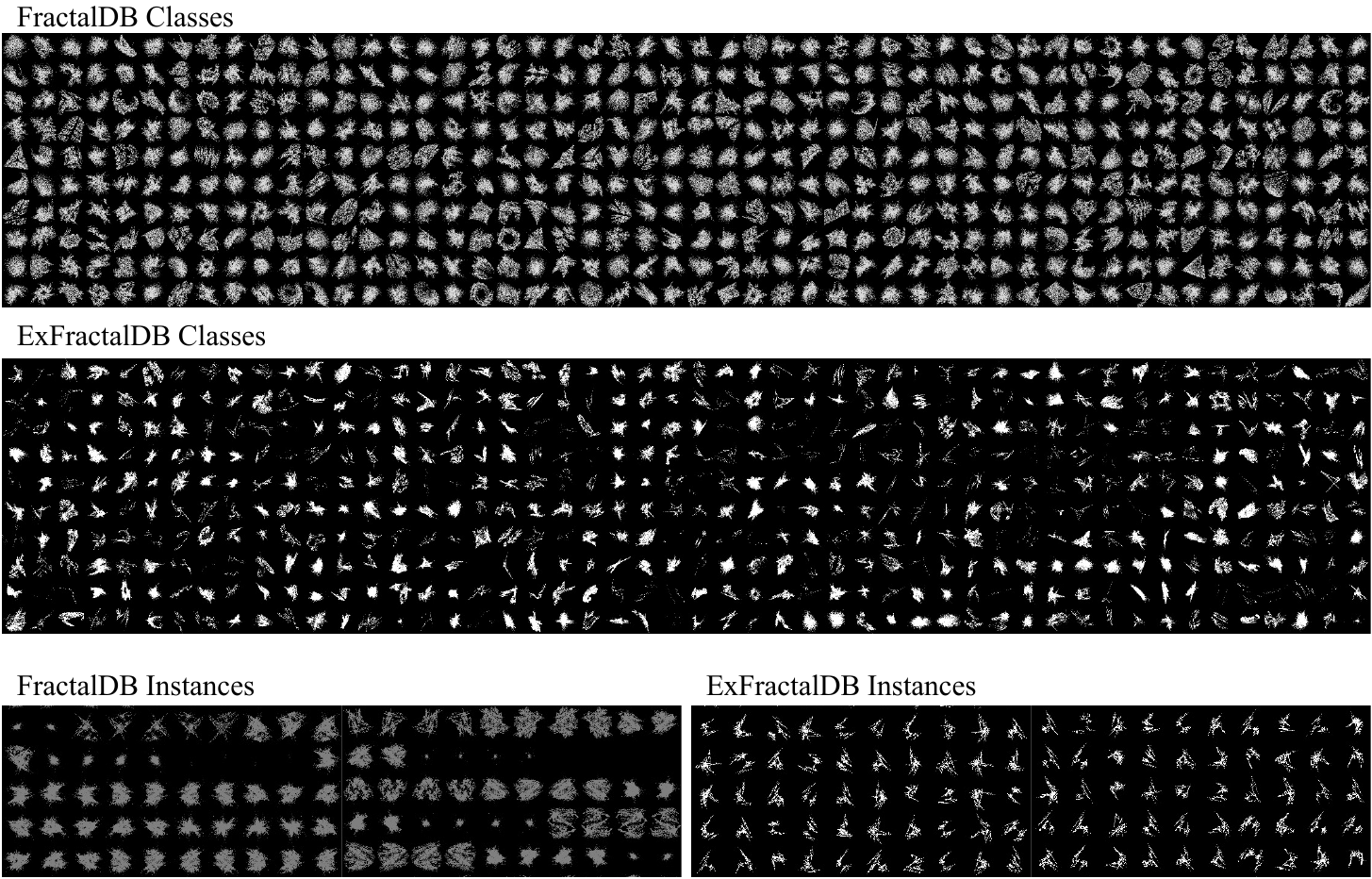}
  \caption{Classes and instances in FractalDB and ExFractalDB.}
  \label{fig:fractals_cat_ins}
\end{figure*}

\begin{figure*}[h]
  \centering
  \includegraphics[width=0.9\linewidth]{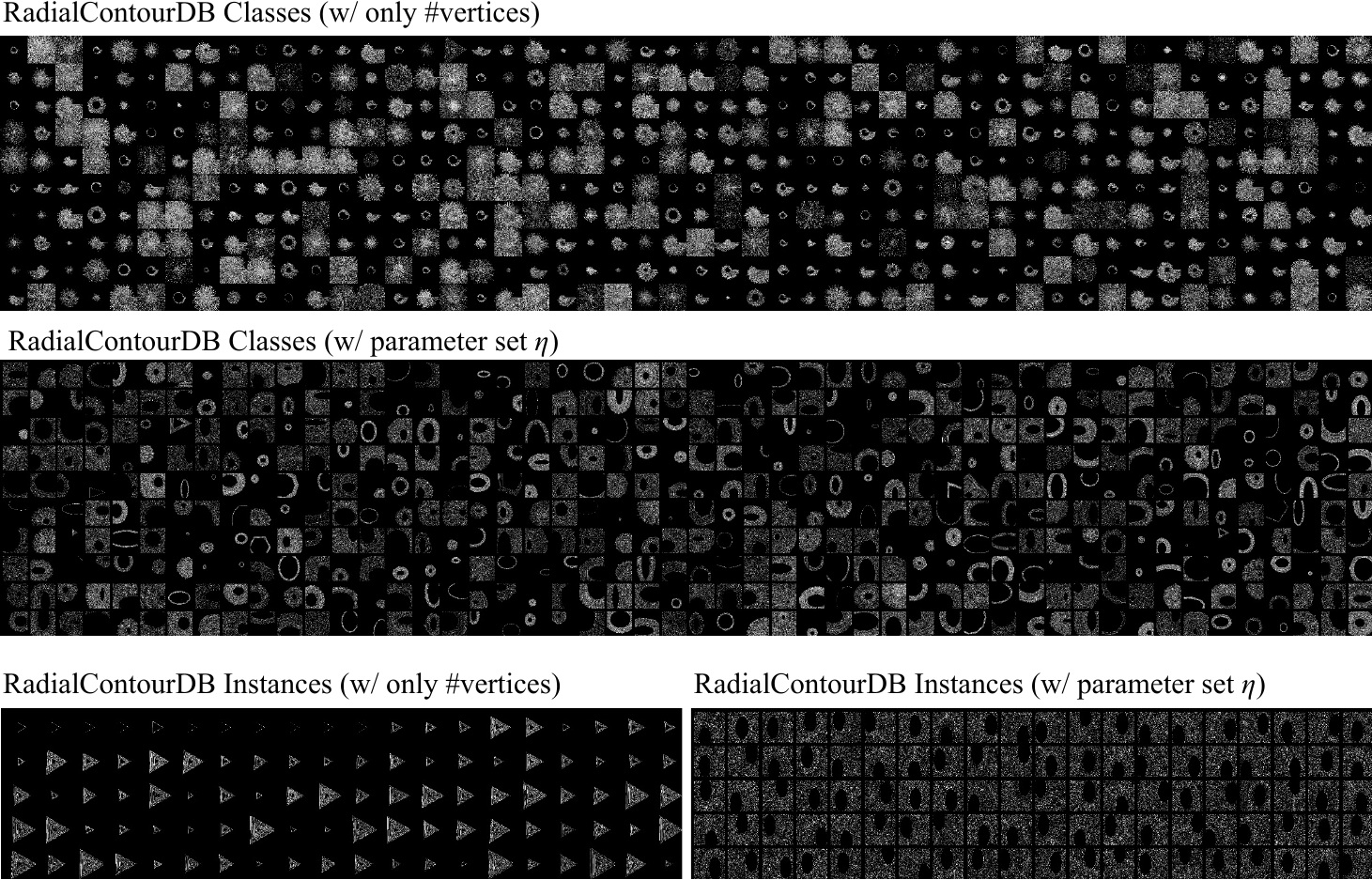}
  \caption{Classes and instances in RCDB with only \#vertices and parameter set $\eta$.}
  \label{fig:rcdb_cat_ins}
\end{figure*}

\begin{figure*}[h]
  \centering
  \includegraphics[width=0.80\linewidth]{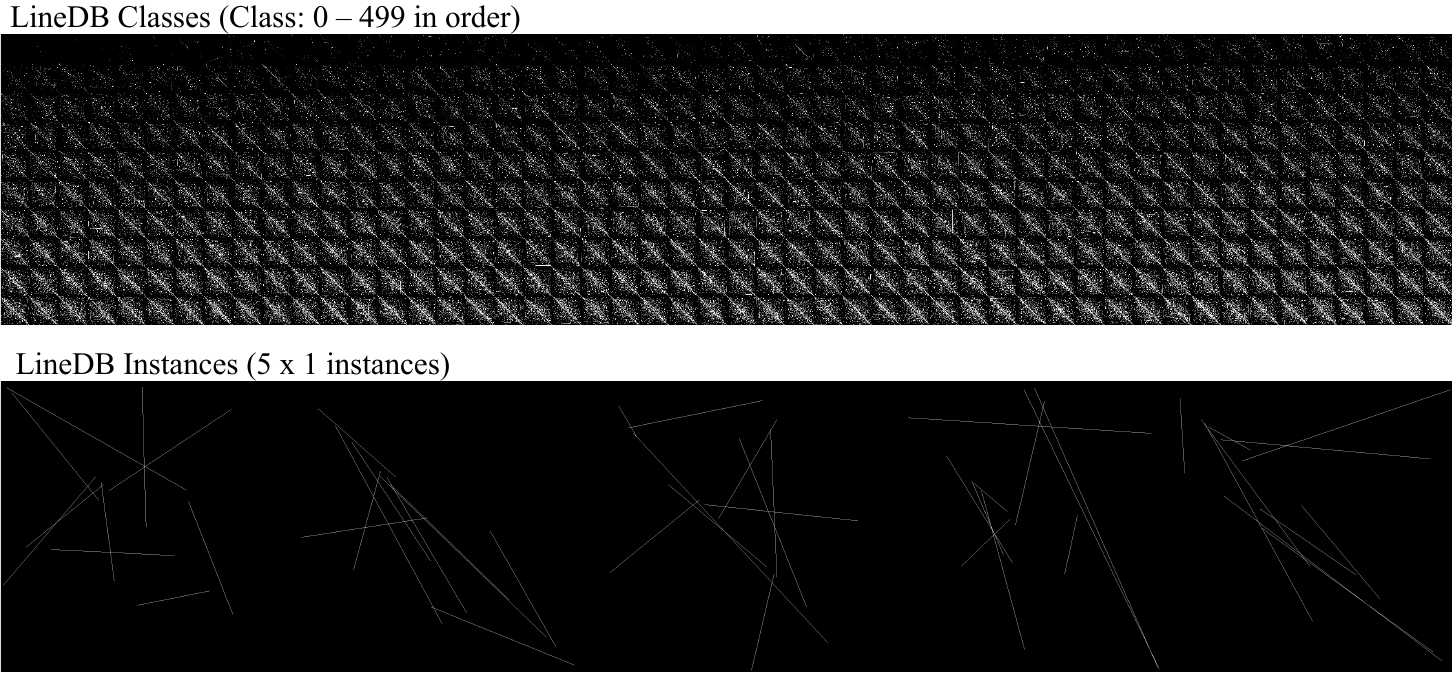}
  \caption{Classes and instances in LineDB.}
  \label{fig:linedb_cat_ins}
\end{figure*}

ExFractalDB consists of $C=$1k classes.
MV-FractalDB generates 12 images from \textit{fixed} viewpoints, whereas ExFractalDB \textit{randomly} selects and projects 2D images from 3D models.

\noindent \textbf{Scaling ExFractalDB.}
The potential of ViTs can only be realized through pre-training on huge datasets with real images. However, previous work used FDSL models pre-trained on only relatively small datasets (on the order of 1M to 10M images)~\cite{NakashimaarXiv2021}. In the present work, we increased the size of the dataset to 21M and 50M images by simply increasing the number of classes, which is a trivial task for FDSL. This resulted in datasets with $C = $ 21k and 50k classes, respectively.
For each class, 25 3D fractal instances were generated.
To increase the variation of the projected images, each instance was captured by a virtual camera from 40 positions, which were randomly and uniformly chosen from the surface of a unit sphere. As a result, $25$ instances $\times$ $40$ viewpoints $= 1,000$ images were generated for each class.

\section{Details in FDSL datasets}
\label{sec:dataset_details_in_fdsl}

Here, we list and describe the FDSL datasets used in our experiments. Table~\ref{tab:fdsl_class_instance} explains the relationships between classes and instances in FDSL datasets for representative FractalDB, RCDB, Dead Leaves, BezierCurveDB, LineDB. Some of the important datasets include the visual examples in Figures~\ref{fig:fractals_cat_ins}, \ref{fig:rcdb_cat_ins}, and \ref{fig:linedb_cat_ins}.

\textbf{Extended FractalDB (ExFractalDB).}
We list example images of 500 randomly selected classes with 100 instances per class in Figure~\ref{fig:fractals_cat_ins}. The figure shows images of both the FractalDB and ExFractalDB. We also describe the definitions of classes and instances in FractalDB, MV-FractalDB, and ExFractalDB for the preliminary study and experiments in Table~\ref{tab:fdsl_class_instance}. 

\textbf{RadialContourDB (RCDB).} We show example images of 500 randomly selected classes with 100 instances per class in Figure~\ref{fig:rcdb_cat_ins}. The figure shows RCDB images with the class definition of only \#vertices and parameter set $\eta$. We also describe the definitions of classes and instances of both RCDB for the experiments shown in Table~\ref{tab:fdsl_class_instance}.

\textbf{Dead Leaves.} 
Originally, Dead Leaves~\cite{RudermanVisionResearch97_deadleaves} was a the combination of simple patterns such as circles, triangles, and squares. Baradad \textit{et al.}~\cite{Baradad2021_deadleaves} assigned labels through SSL. However, in our case, we labeled the dataset using the FDSL framework. We simply used the ratios of circles, triangles, and squares in the images, and use those as labels. The accuracy of the \{C10, C100, Cars, Flowers\} datasets changed from \{89.8, 71.4, 32.0, 91.6\} when trained with SimCLRV2 to \{95.9, 79.6, 72.8, 96.9\} when pre-trained with FDSL.
The definitions of classes and instances are shown in Table~\ref{tab:fdsl_class_instance}.

\begin{table}[t]
    \begin{center}
    \caption{Effect of batch size on ExFractalDB / RCDB. Best values for each dataset are in bold.}
    \begin{tabular}{lccccc} \toprule[0.8pt]
        Batch size &  C10 & C100 & Cars & Flowers \\\midrule[0.5pt]
        32 & 96.5 / 94.9 & 79.2 / 77.4 & 82.8 / 68.4 & 96.4 / 95.6 \\
        64 & 97.2 / 96.7 & 81.2 / 79.5 & 87.7 / 81.7 & 98.2 / 97.8 \\
        128 & \textbf{97.4} / 96.7 & 81.9 / 79.5 & 87.8 / 81.7 & 97.8 / 97.9 \\
        256 & 97.1 / 96.8 & 80.9 / 81.2 & 89.0 / \textbf{85.5} & 98.2 / \textbf{98.5} \\
        512 & 97.2 / 96.9 & 81.8 / \textbf{82.1} & \textbf{90.0} / 84.9 & \textbf{99.4} / \textbf{98.5} \\
        1024 & 97.3 / \textbf{97.0} & \textbf{82.6} / 81.8 & 89.9 / \textbf{85.5} & \textbf{99.4} / 98.2 \\
        2048 & 96.9 / \textbf{97.0} & 81.8 / 81.7 & 88.2 / 84.8 & 99.1 / 98.2 \\
        4096 & 96.6 / 96.3 & 81.4 / 81.6 & 86.7 / 84.6 & 98.6 / 98.0 \\
        \bottomrule[0.8pt]
    \end{tabular}
    \label{tab:batch_size}
    \end{center}
\end{table}

\begin{table}[t]
    \begin{center}
    \caption{Effect of pre-training epoch on ExFractalDB / RCDB. Best values for each dataset are in bold.}
    \begin{tabular}{lccccc} \toprule[0.8pt]
        Batch size &  C10 & C100 & Cars & Flowers \\\midrule[0.5pt]
        300 & \textbf{97.3} / \textbf{97.0} & \textbf{82.6} / 81.8 & 89.9 / \textbf{85.5} & \textbf{99.4} / 98.2 \\
        500 & \textbf{97.3} / \textbf{97.0} & 81.6 / \textbf{81.9} & \textbf{90.1} / \textbf{85.5} & 99.3 / \textbf{98.4} \\
        \bottomrule[0.8pt]
    \end{tabular}
    \label{tab:training_epoch}
    \end{center}
\end{table}

\textbf{BezierCurveDB.}
The previous BezierCurveDB had only \#lines and \#dots as formula-supervision parameters. 
However, as a consequence of hypothesis 2, we improved the Bezier Curves by adding a restriction to the control point coordinates to increase the number of formula-supervision parameters. In addition to the conventional instance augmentations, we augmented the instances further by randomly selecting vectors with connecting control points.
The definitions of classes and instances are shown in Table~\ref{tab:fdsl_class_instance}.

\textbf{LineDB.} We list example images of 500 randomly selected classes with 5 instances per class in Figure~\ref{fig:linedb_cat_ins}. The class definition is the number of lines. Therefore, 1,000 classes are divided from 0 to 999 lines in an image. In the instance definition, there are randomly generated lines in each image. The definition of classes and instances are also shown in Table~\ref{tab:fdsl_class_instance}.

\section{Experiments}

\subsection{Hyperparameter tuning}
\label{sec:hyperparameter_tuning}

At the beginning of the experiments in our study, we found batch size and pre-training epoch to be important hyperparameters.

\textbf{Effect of batch size (Table~\ref{tab:batch_size}).}
We show the effect of batch size for FDSL in Table~\ref{tab:batch_size}. As shown, the optimal batch size for FDSL is around 1,024. However, even for a batch size of 64, which is smaller than the batch size of a real-image dataset (e.g., with CNN, SSL and ImageNet in~\cite{ChenCVPR2021_simsiam}), the scores do not significantly drop from their highest values. This suggests that FDSL has a smaller optimal batch size compared to that of SSL with real images. Hereafter, we set the batch size to 1,024 in our experiments.

\textbf{Effect of pre-training epochs (Table~\ref{tab:training_epoch}).}
We show the effect of pre-training epochs for FDSL in Table~\ref{tab:training_epoch}. Due to limited computational resources, we implemented only 300 epochs as a basic training length and 500 epochs for a longer training length. As shown in the table, the pre-trained model with a longer 500-epoch pre-training is slightly better than that of the pre-trained model with basic 300-epoch pre-training. Therefore, we used 300 epochs and 500 epochs for comparison with other related methods.

\subsection{Verification of hypotheses 1 and 2}
\label{sec:exploration_study}
In the verification, we chose the same datasets for fine-tuning as those in previous studies~\cite{TouvronICML2021,NakashimaarXiv2021}, namely CIFAR-10/100 (C10/C100)~\cite{Krizhevsky2009_cifar}, Stanford Cars (Cars)~\cite{Krause3DRR2013_cars}, and Flowers~\cite{Nilsback08_flowers}.
We used the same hyperparameters and data augmentation as those in \cite{TouvronICML2021}, except for the batch size and pre-training epochs discussed in Section~\ref{sec:hyperparameter_tuning}.
FractalDB in these experiments has 1k classes, each with 1k instances.
We updated all layers during the pre-training and fine-tuning for all experiments.
The following tables and their descriptions correspond to hypothesis 1 or 2.

% Comparison of FDSL methods
\noindent \textbf{Hypothesis 1: Comparison of FDSL methods (Table~\ref{tab:fdsl_labels}).}
We compared various types of mathematically generated dataset.
Perlin noise and Bezier curves are from~\cite{KataokaACCV2020}.

The results show that pre-training with RCDB and FractalDB give the highest fine-tuning accuracy, closely followed by BezierCurveDB. In RCDB, we only changed the number of vertices ($n$ in parameter set $\eta$).
Regarding hypothesis 1, we confirmed that image representation using object contours tends to yield higher scores.

\begin{table}[t]
    \begin{center}
    \caption{Comparison of FDSL methods. Hereafter, best values are in bold.}
    \begin{tabular}{lccccc} \toprule[0.8pt]
        Pre-training &  C10 & C100 & Cars & Flowers \\\midrule[0.5pt]
        From scratch & 78.3 & 57.7 & 11.6 & 77.1 \\ 
        PerlinNoiseDB~\cite{KataokaACCV2020} & 95.0 & 78.4 & 70.6 & 96.1 \\ 
        Dead Leaves~\cite{Baradad2021_deadleaves} & 95.9 & 79.6 & 72.8 & 96.9 \\ 
        \rowcolor[gray]{0.8} BezierCurveDB~\cite{KataokaACCV2020} & 96.7 & 80.3 & 82.8 & 98.5 \\
        \rowcolor[gray]{0.8} RCDB & \textbf{96.8} & \textbf{81.6} & 84.2 & \textbf{98.7} \\
        \rowcolor[gray]{0.8} FractalDB~\cite{NakashimaarXiv2021} & \textbf{96.8} & \textbf{81.6} & \textbf{86.0} & 98.3 \\\midrule[0.5pt]
        \begin{minipage}{20mm}
        \centering
        \includegraphics[width=6.8cm]{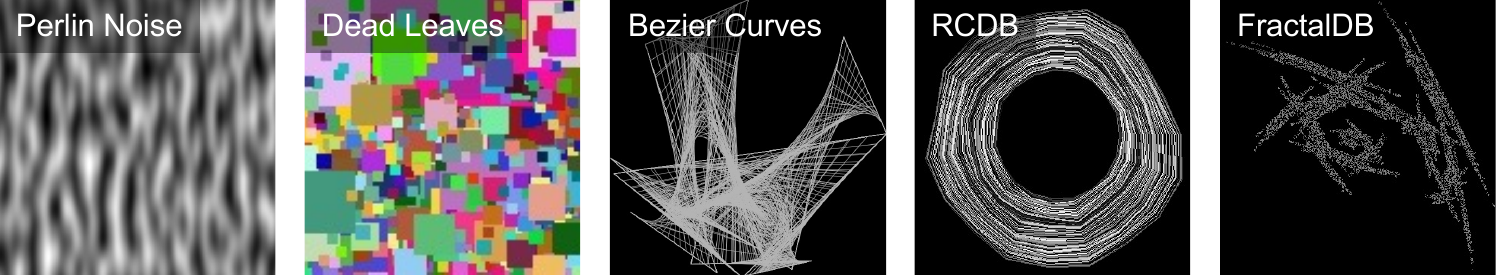}
        \end{minipage} \\
        \bottomrule[0.8pt]
    \end{tabular}
    \label{tab:fdsl_labels}
    \end{center}
\end{table}

\begin{table}[t]
    \begin{center}
    \caption{Relationship between \#vertices and accuracy in RCDB.}
    \begin{tabular}{lccccc} \toprule[0.8pt]
        Range & \#Vertices &  C10 & C100 & Cars & Flowers \\\midrule[0.5pt]
        \rowcolor[gray]{0.8} 3--102 & 100 & \textbf{95.5} & \textbf{79.4} & \textbf{78.4} & \textbf{96.4} \\
        103--202 & 100 & 94.2 & 76.3 & 55.8 & 95.9 \\
        203--302 & 100 & 71.3 & 46.9 & 4.9 & 49.8 \\
        303--402 & 100 & 59.4 & 33.9 & 2.5 & 26.8 \\
        403--502 & 100 & 40.1 & 13.6 & 0.8 & 5.3 \\\midrule[0.5pt]
        \rowcolor[gray]{0.8} 3--202 & 200 & \textbf{96.5} & \textbf{80.2} & \textbf{80.8} & \textbf{98.1} \\
        303--502 & 200 & 93.6 & 76.1 & 44.8 & 94.1 \\\midrule[0.5pt]
        \rowcolor[gray]{0.8} 3--502 & 500 & \textbf{96.4} & \textbf{80.7} & \textbf{83.0} & \textbf{98.5} \\\midrule[0.5pt]
        \begin{minipage}{20mm}
        \centering
        \includegraphics[width=8.0cm]{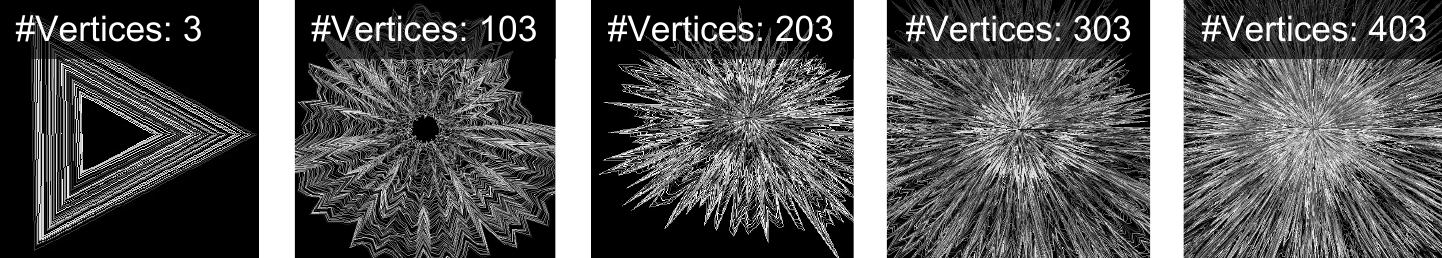}
        \end{minipage} \\
        \bottomrule[0.8pt]
    \end{tabular}
    \label{tab:numof_vertices}
    \end{center}
\end{table}

% Complexity of object contours in RCDB
\noindent \textbf{Hypothesis 1: Complexity of object contours in RCDB (Table~\ref{tab:numof_vertices}).}
The complexity of the object contours in RCDB can be controlled by changing the number of vertices.
We split the classes depending on the number of vertices.

Table~\ref{tab:numof_vertices} shows the results of RCDB for various ranges of categories.
A comparison of the results obtained with 3--502 vertices (500 classes), 3--202 vertices (200 classes), and 3--102 vertices (100 classes) indicates that there exists an optimal degree of contour complexity.
For RCDB pre-trained ViTs with 100 classes, the best results are obtained with 103--202 vertices (100 classes); accuracy saturates at about 203--302 vertices (100 classes).
Pre-training on RCDB with 203--302, 303--402, and 403--502 vertices led to lower scores than those for training from scratch (Table~\ref{tab:comparison}).
This means that overly complicated object contours inhibit the acquisition of visual representation during the pre-training phase.
Therefore, the merged dataset with 3--202 (200 classes) is almost enough to pre-train a ViT that performs with accuracy similar to that of the pre-trained model with 3--502 vertices (500 classes). Here, the other 300 classes (303--502) help to slightly improve the pre-training effect. We basically assigned the 3--502 as a range of vertices in RCDB.

% Increased number of parameters in FDSL pre-training
\noindent \textbf{Hypothesis 2: Increasing the number of parameters in FDSL pre-training (Table~\ref{tab:multiple_parameters}).}
We increased the number of parameters to create FDSL labels.
The results for BezierCurveDB, RCDB, and ExFractalDB are shown in Table~\ref{tab:multiple_parameters}.

\begin{table}[t]
    \begin{center}
    \caption{Effect of increased parameters on FDSL methods. Values in parentheses indicate the difference from the case with fewer parameters.}
    \begin{tabular}{lccccc} \toprule[0.8pt]
        Pre-training &  C10 & C100 & Cars & Flowers \\\midrule[0.5pt]
        %DL & 95.1 (-0.8) & 79.0 (-0.6) & 59.7 (-13.1) & 95.7 (-1.2) \\ 
        BezierCurveDB & 96.9 \scriptsize{(0.2)} & 81.4 \scriptsize{(1.1)} & 85.9 \scriptsize{(3.1)} & 97.9 \scriptsize{(-0.6)} \\
        \rowcolor[gray]{0.8} RCDB & 97.0 \scriptsize{(0.2)} & \textbf{82.2} \scriptsize{(0.6)} & 86.5 \scriptsize{(2.4)} & \textbf{98.9} \scriptsize{(0.2)} \\
        \rowcolor[gray]{0.8} ExFractalDB & \textbf{97.2} \scriptsize{(0.4)} & 81.8 \scriptsize{(0.2)} & \textbf{87.0} \scriptsize{(1.0)} & \textbf{98.9} \scriptsize{(0.6)} \\
        \bottomrule[0.8pt]
    \end{tabular}
    \label{tab:multiple_parameters}
    \end{center}
\end{table}

\begin{figure*}[t]
\centering
%polygon
%\subfigure[\#contours ($N$)]
\subfigure[$N$]{\includegraphics[width=0.18\linewidth]{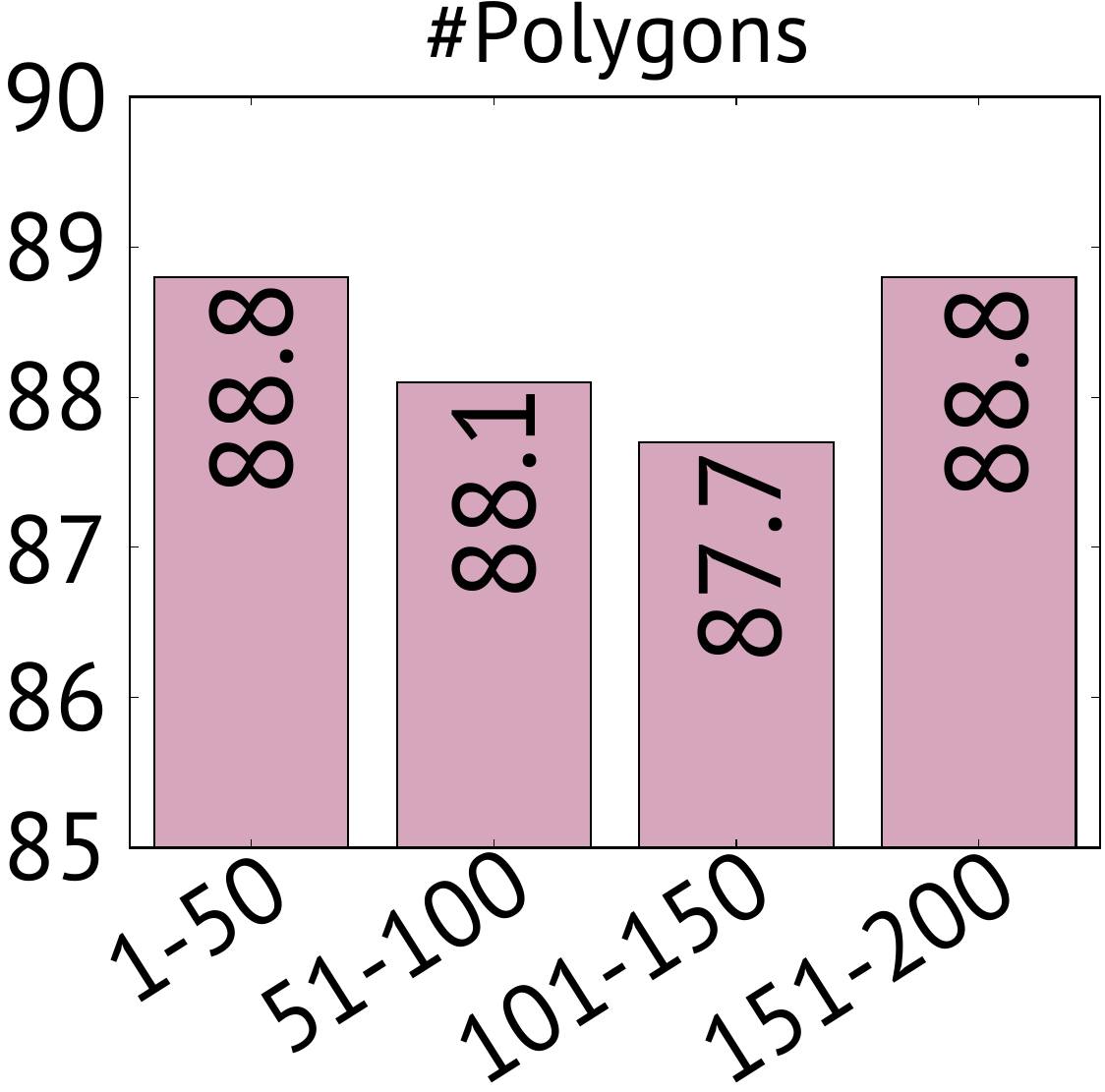}
\label{fig:polygon}}
%radius
%\subfigure[Radius ($r$)]
\subfigure[$r$]{\includegraphics[width=0.18\linewidth]{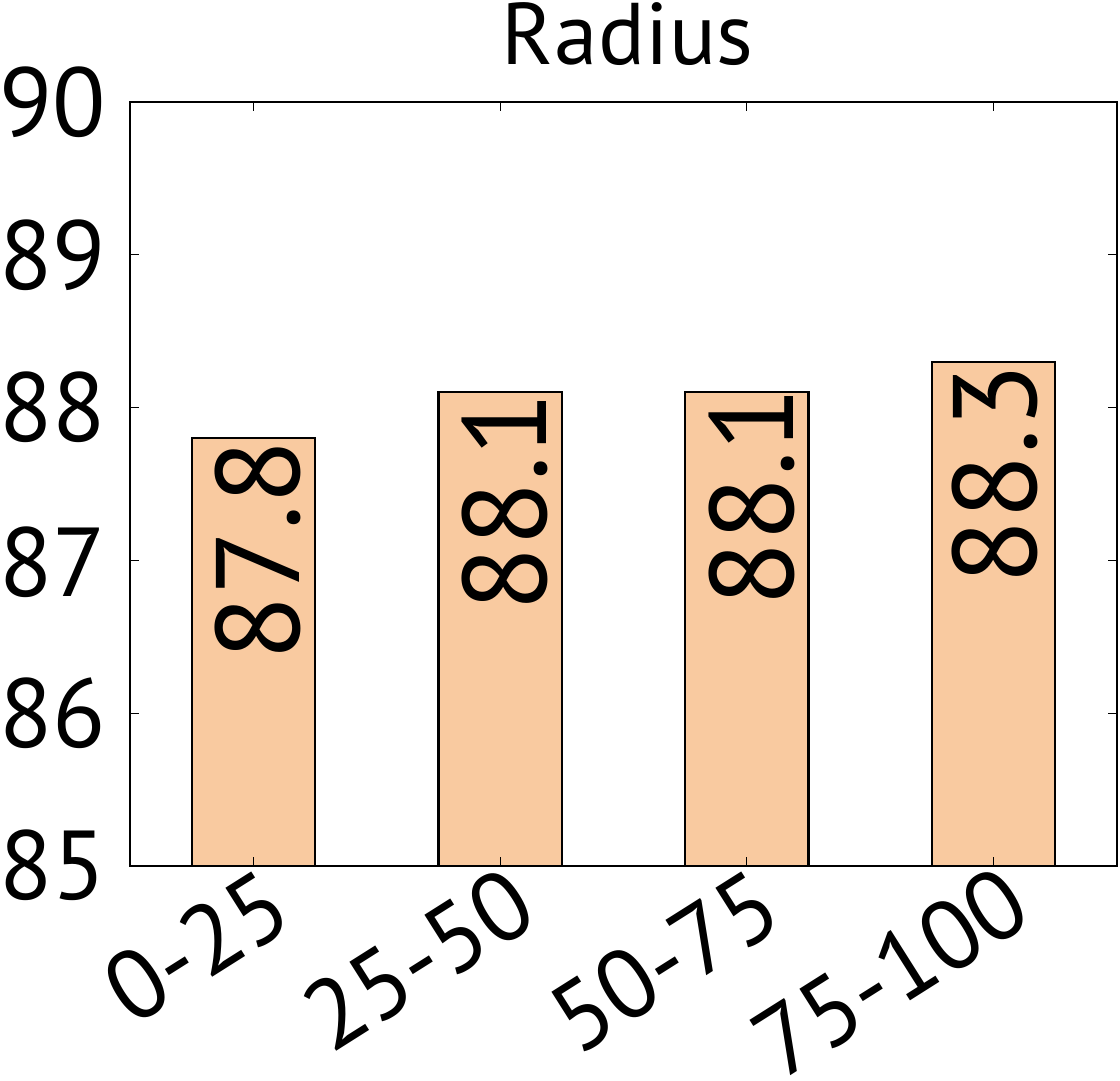}
\label{fig:radius}}
%width
%\subfigure[Polygon width ($l_{w}$)]
\subfigure[$l_{w}$]{\includegraphics[width=0.18\linewidth]{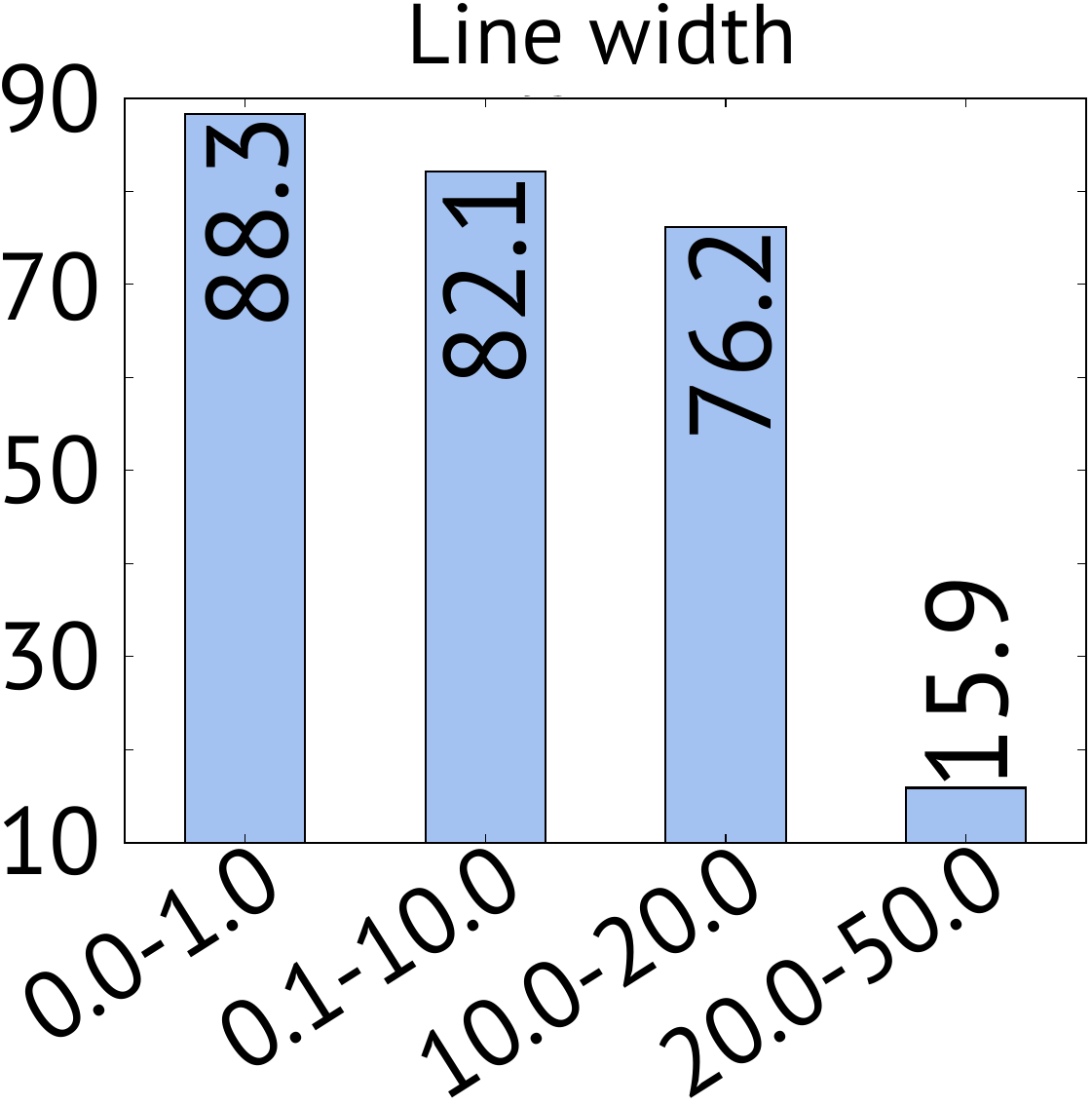}
\label{fig:width}}
%resizing factor
%\subfigure[Resizing factor (${\bf o}$)]
\subfigure[${\bf o}$]{\includegraphics[width=0.18\linewidth]{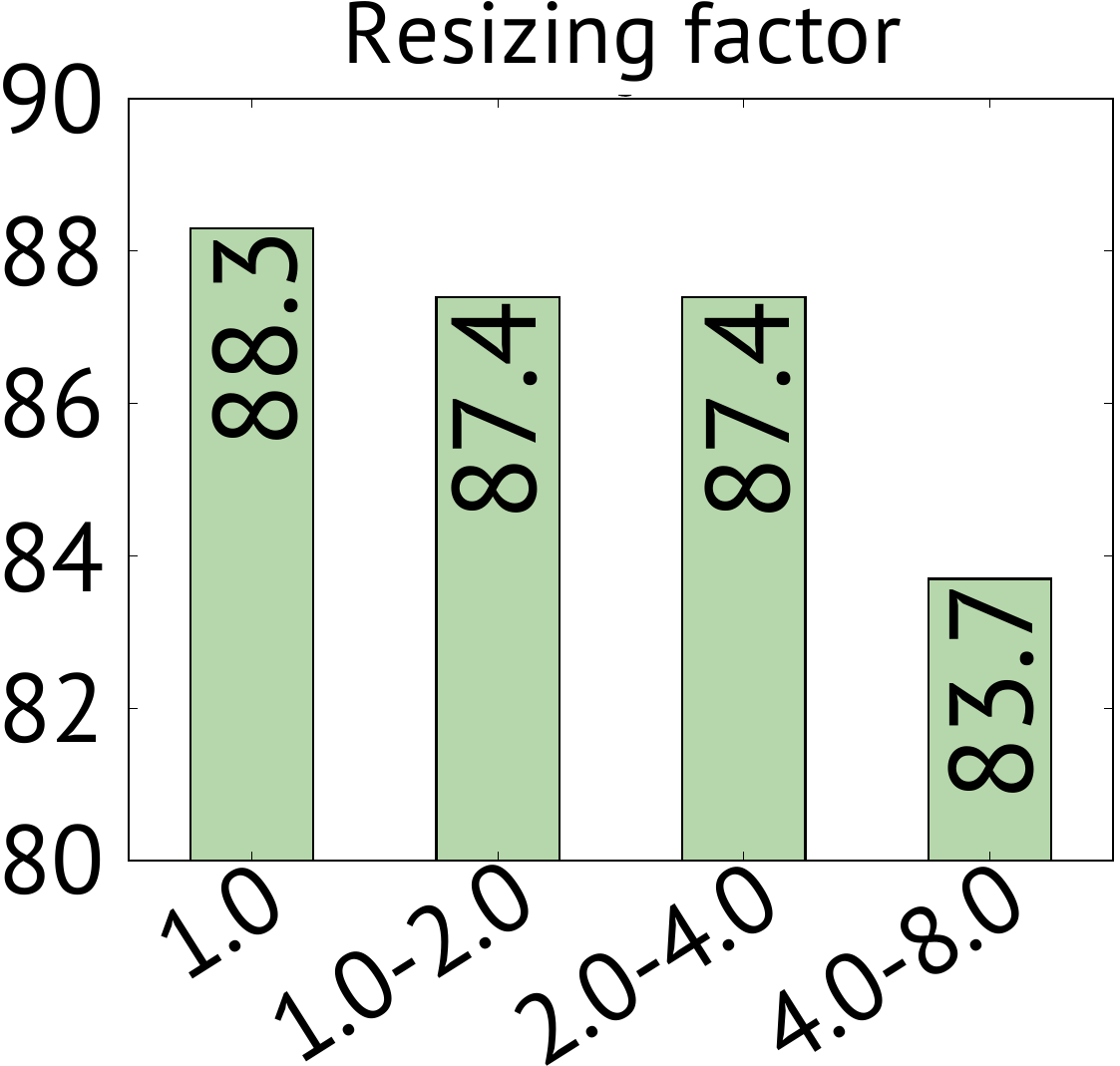}
\label{fig:resizing_factor}}
%perlin
%\subfigure[Perlin noise (${\bf \lambda}$)] 
\subfigure[${\bf \lambda}$]{\includegraphics[width=0.18\linewidth]{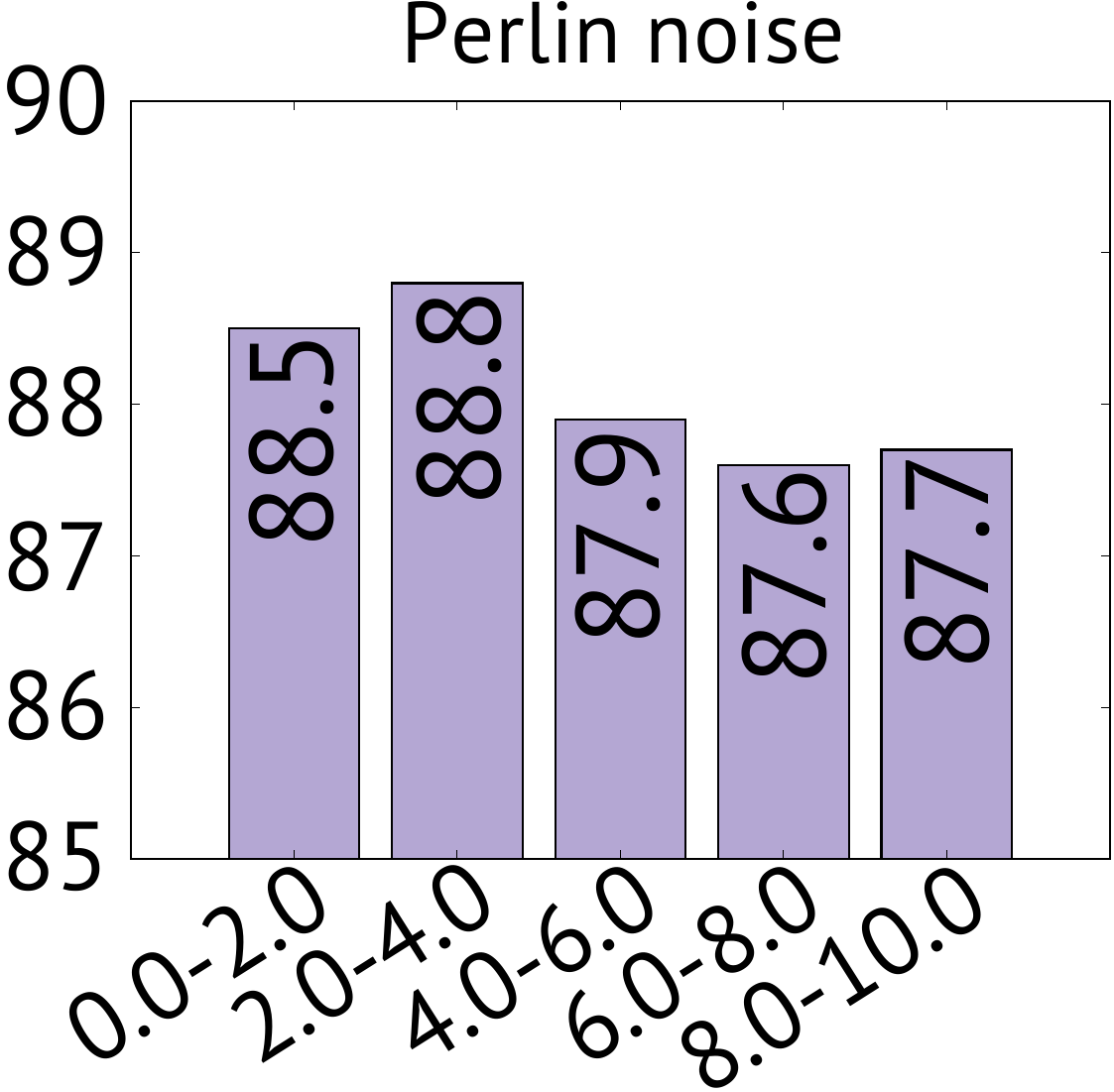}
\label{fig:perlin_noise}}
\caption{Parameter tuning for RCDB. Tuning was conducted with C10, C100, Cars, and Flowers. The values in the graphs show the average rates for the four datasets.}
\label{fig:rcdb_exploration}
\end{figure*}

\begin{table}[t]
    \begin{center}
    \caption{Comparisons of class definition type (from 2D IFS and 3D IFS) and instance augmentation (MV-FractalDB: fixed viewpoints and ExFractalDB: random viewpoints).} % Instance augmentation on ExFractalDB. Best values are in bold.
    \begin{tabular}{lccccc} \toprule[0.8pt]
        Pre-training & IFS & C10 & C100 & Cars & Flowers \\\midrule[0.5pt]
        FractalDB~\cite{KataokaACCV2020} & 2D & 96.8 & 81.6 & 86.0 & 98.3 \\
        MV-FractalDB~\cite{YamadaIROS2021} & 3D & 96.9 & 81.4 & 86.5 & 98.5 \\
        \rowcolor[gray]{0.8} ExFractalDB & 3D & \textbf{97.2} & \textbf{81.8} & \textbf{87.0} & \textbf{98.9} \\
        \bottomrule[0.8pt]
    \end{tabular}
    \label{tab:multiview_fractaldb}
    \end{center}
\end{table}

\begin{figure*}[t]
\centering
%imaganet
\subfigure[C10] {\includegraphics[width=0.18\linewidth]{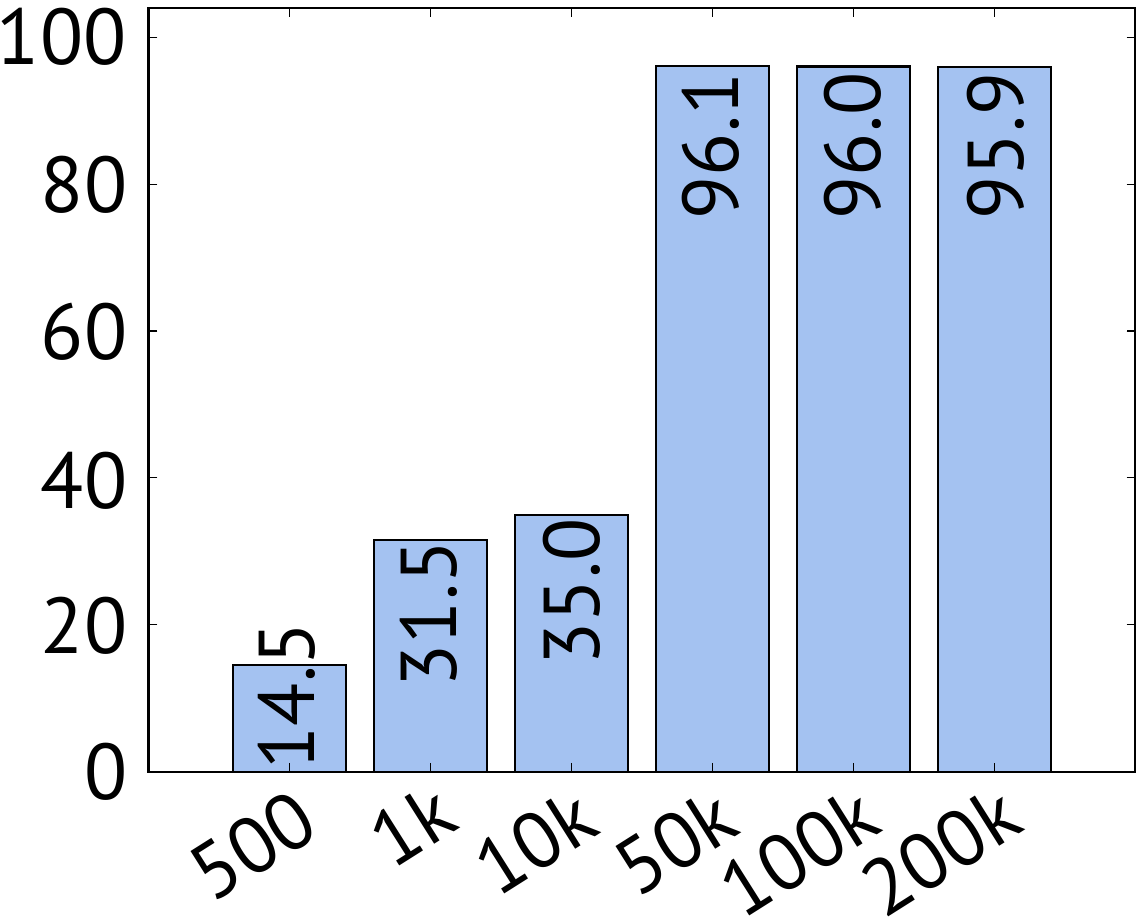}
\label{fig:point_fractal_cifar10}}
%places
\subfigure[C100] {\includegraphics[width=0.18\linewidth]{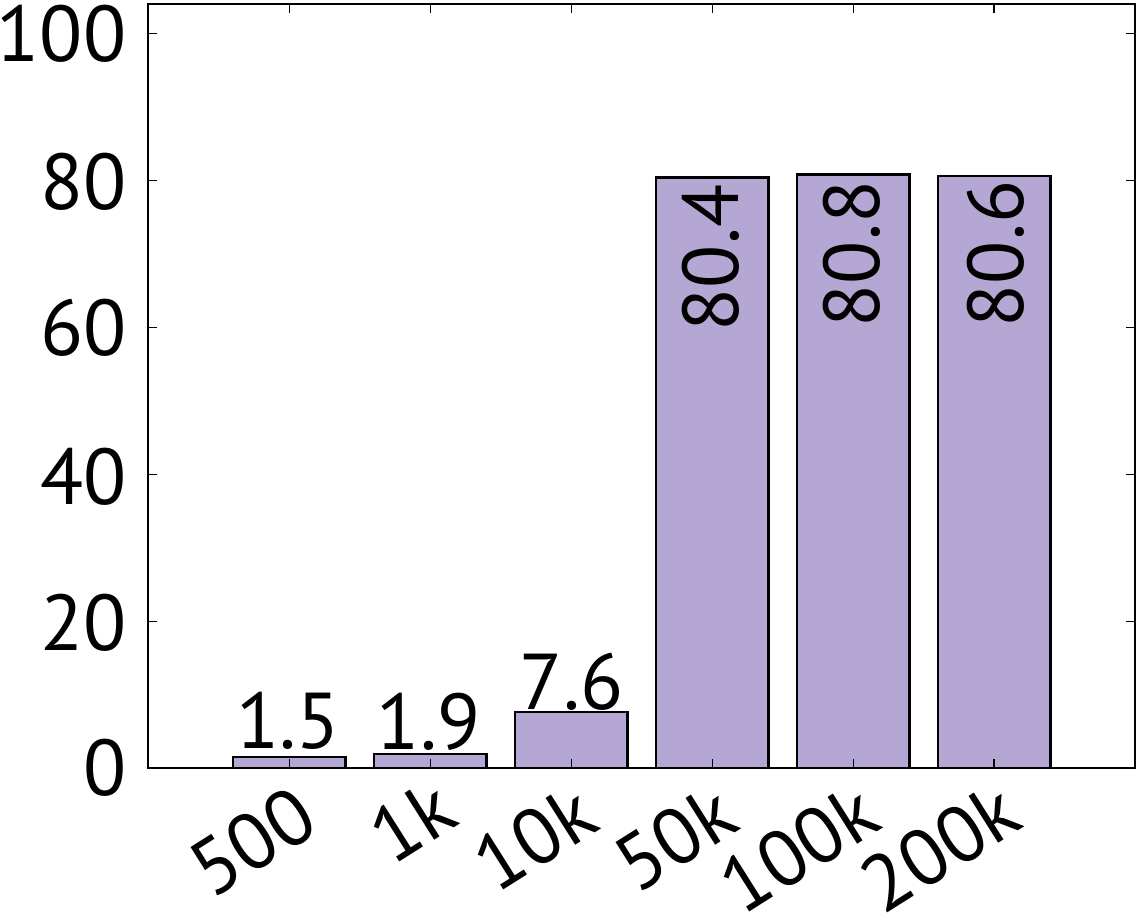}
\label{fig:point_fractal_cifar100}}
%fractaldb baseline
\subfigure[Cars]{\includegraphics[width=0.18\linewidth]{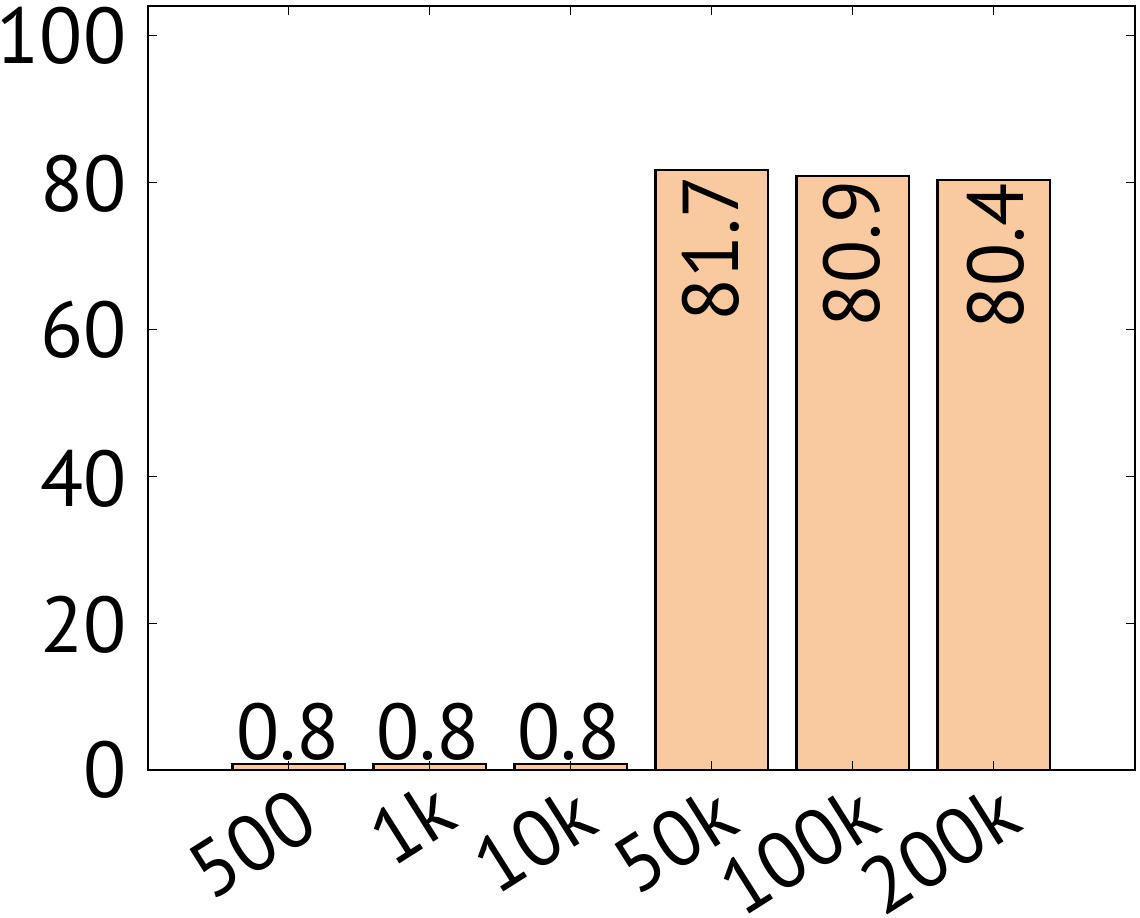}
\label{fig:point_fractal_cars}}
%fractaldb (color random)
\subfigure[Flowers]{\includegraphics[width=0.18\linewidth]{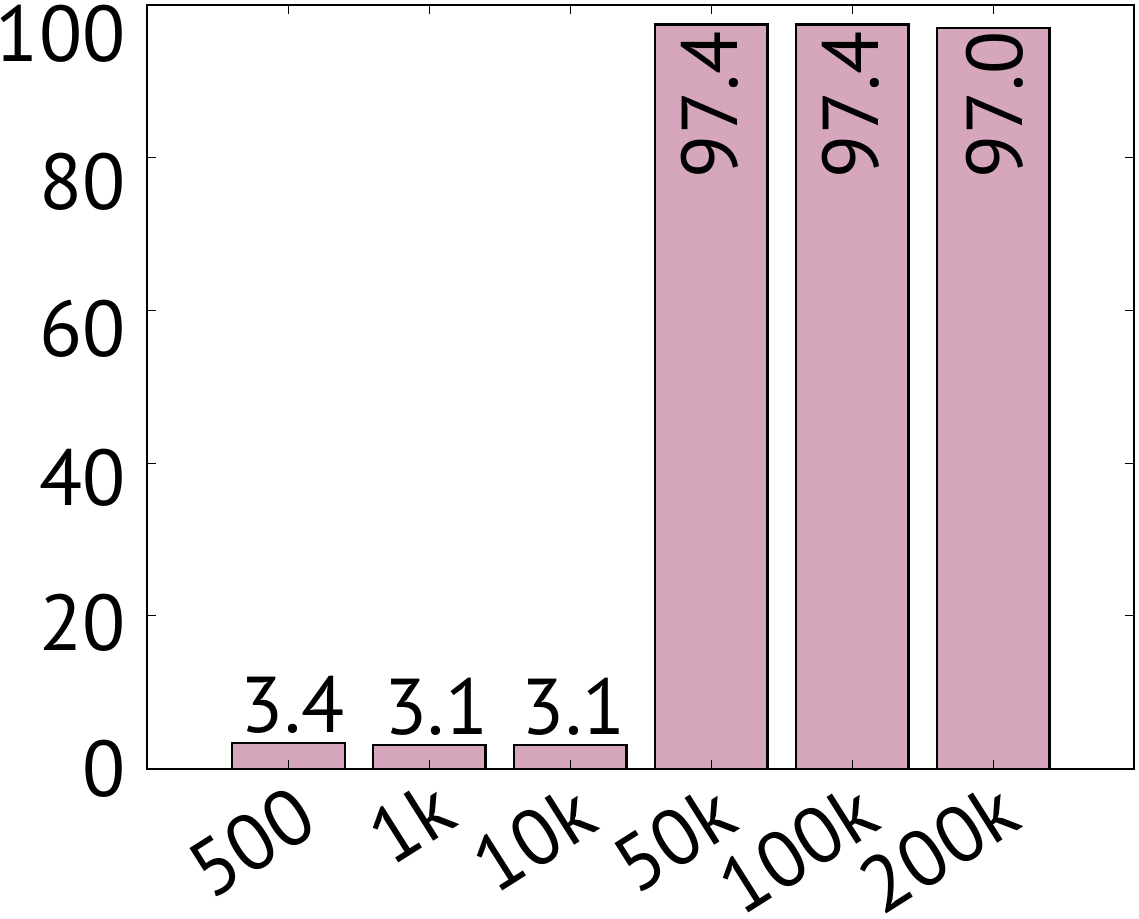}
\label{fig:point_fractal_flowers}}
%rcdb
\subfigure[RCDB w/ and w/o corruption]{\includegraphics[width=0.20\linewidth]{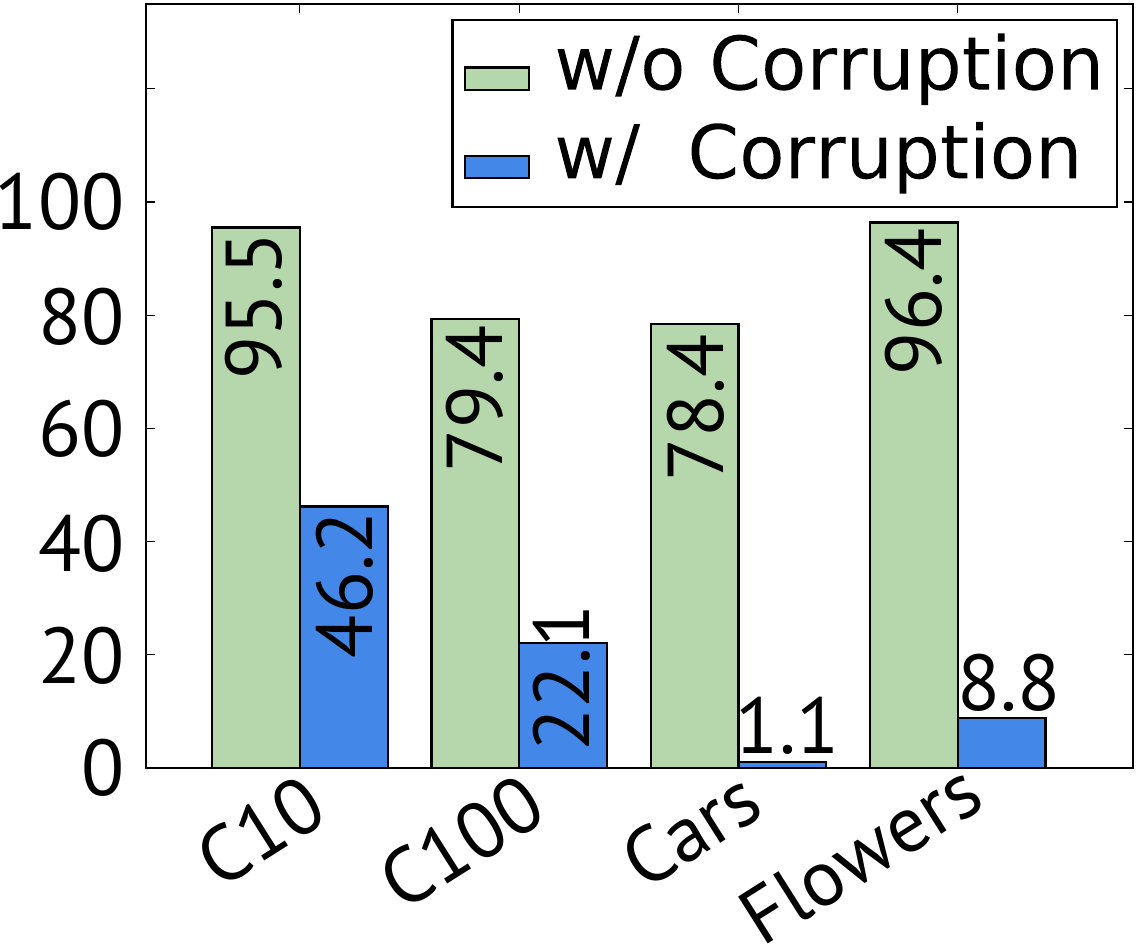}
\label{fig:corruption_rcdb}}
\\
% #Points and fractal images
\subfigure[(Left, center-left) Point rendering in FractalDB-1k with 10k and 50k points. We executed pre-training with 500, 1k, 10k, 50k, 100k, and 200k points in FractalDB-1k. (Center-right, right) The attention maps with the pre-trained models on FractalDB with 10k and 50k points, respectively.]{\includegraphics[width=0.50\linewidth]{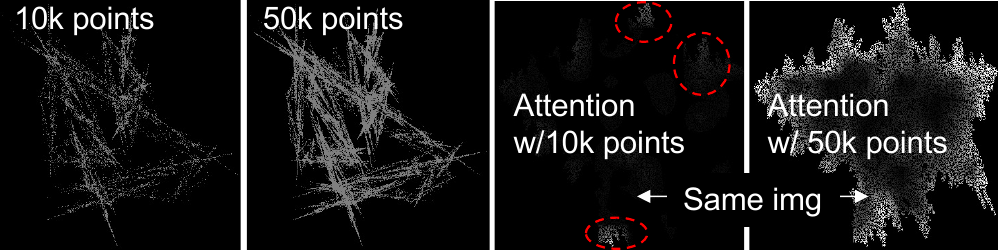}
\label{fig:point_fractal_images}}
\hspace{10pt}
% RCDB image with corruption
\subfigure[(Left) Example of RCDB with broken contours. We deliberately drew 1k lines with the same color as the background. (Center, right) Attention maps with the pre-trained models on RCDB with and without broken object contours, respectively.]{\includegraphics[width=0.40\linewidth]{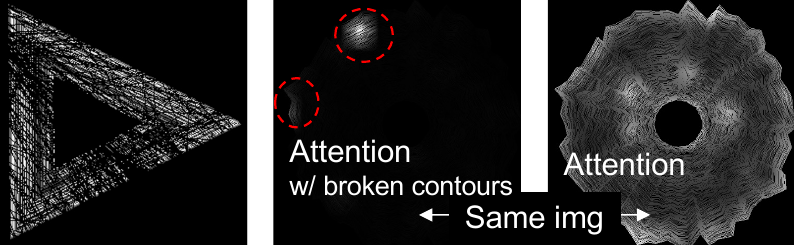}
\label{fig:rcdb_image_corruption}}
%\caption{Results and image examples in point-rendered FractalDB-1k pre-training.}
\caption{Results, image examples, and attention maps in point-rendered FractalDB-1k (a, b, c, d, f) and RCDB with corruption (e, g). Although the fractal images with 50k (or more) points and radial contours successfully trained the models for visual representation, the fractal images with 10k (or fewer) points and radial contours with corruption failed.}
\label{fig:result_fractal_contour}
\end{figure*}

\begin{figure}[h]
  \centering
  \includegraphics[width=0.75\linewidth]{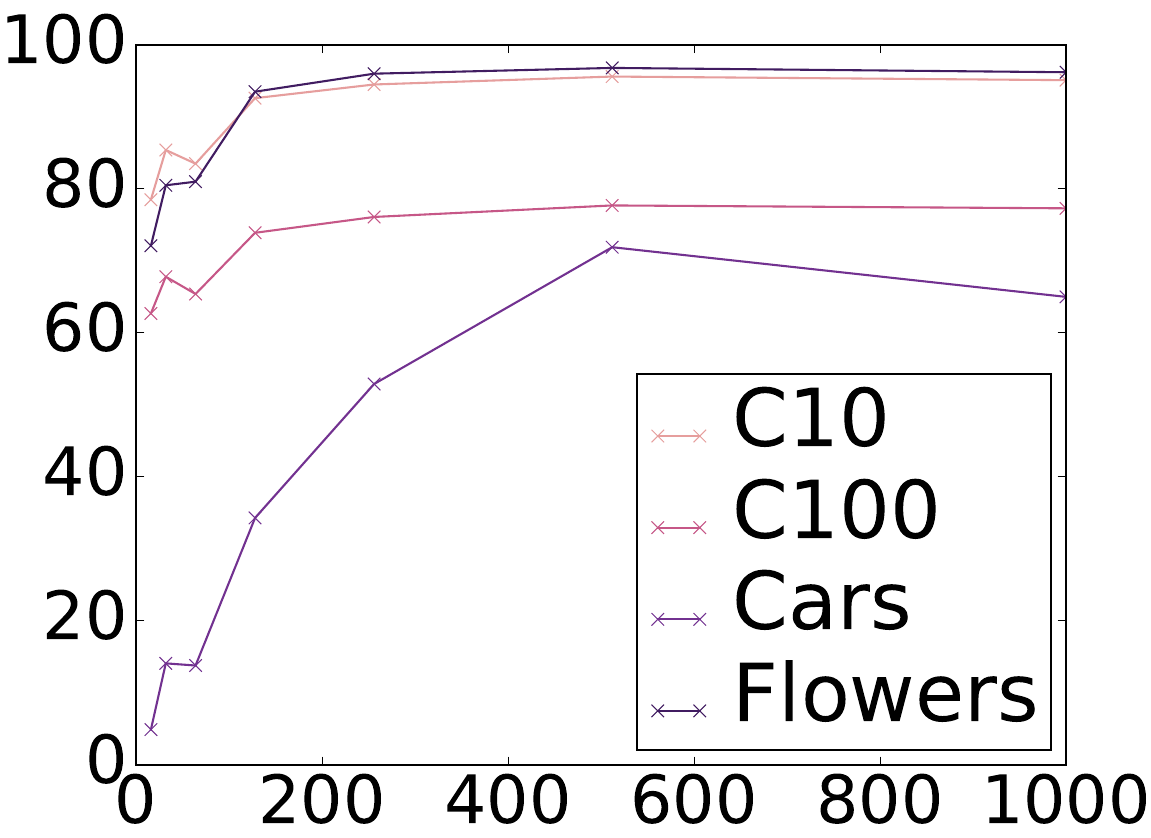}
  \caption{Pre-training with line counting. The figure shows the effect of \#categories on accuracy.}
  \label{fig:linedb}
\end{figure}

% Parameter search for RCDB
\noindent \textbf{Hypothesis 2: Parameter search for RCDB (Figures~\ref{fig:polygon}--\ref{fig:perlin_noise}).}
We explored the parameter set $\eta$ to combine changes in both these parameters and \#vertices. Figures~\ref{fig:polygon}--\ref{fig:perlin_noise} respectively show the effects of the number of polygons, radius, line width, resizing factor, and Perlin noise on accuracy. The parameters for RCDB in Table~\ref{tab:rcdb} are based on the results of this parameter search.

% Increasing number of instances in ExFractalDB
\noindent \textbf{Hypothesis 2:} Comparison of FractalDB~\cite{KataokaACCV2020}, MV-FractalDB~\cite{YamadaIROS2021}, and ExFractalDB (Table~\ref{tab:multiview_fractaldb}).
ExFractalDB renders 3D fractals and projects them onto a 2D image.
MV-FractalDB uses a fixed perspective when projecting 3D fractals onto 2D images; however, the present work used a random perspective.
MV-FractalDB labels the images according to the perspective in addition to the class of fractals; however, we did not consider the perspective labels.

As shown in the visualization (Figure~\ref{fig:figure1}), acquiring 2D images from a 3D model sharpens attention to multiple locations, which seems to be useful for classification (unlike the attention to contours in Figure~\ref{fig:contour_selfattention}), and improves the accuracy itself.

\subsection{Failure modes of FDSL}
We have shown that pre-training with synthetic images that consist of only simple contours can match the accuracy of real images even at a fairly large scale. We now investigate when and how FDSL can fail.

\noindent \textbf{Minimum number of rendering points.}
We investigated the minimum number of points used in the rendering of fractals in FractalDB. Figure~\ref{fig:result_fractal_contour} shows the results and image examples in the point-rendered FractalDB. According to Figures~\ref{fig:point_fractal_cifar10}--\ref{fig:point_fractal_flowers}, the pre-trained models acquire a good representation when the number of fractal points is 50k or higher, at which point the fractal images start to form a contour (Figure~\ref{fig:point_fractal_images}).

\noindent \textbf{Adding corruption (broken object contours).}
We verified RCDB images with and without broken object contours, as shown in Figure~\ref{fig:rcdb_image_corruption}. We deliberately drew 1k lines with the same color as the background. The lengths and positions of the lines were fully randomized. We adjusted the thickness of the lines to corrupt the object contours of RCDB, but the main frame did not disappear, as in Figure~\ref{fig:rcdb_image_corruption}. From the results in Figure~\ref{fig:corruption_rcdb}, the rates change from \{95.5, 79.4, 78.4, 96.4\} with corruption to \{46.2, 22.1, 1.1, 8.8\} without corruption. These results support hypothesis 1, namely that object contours are an essential component in FDSL datasets.

\noindent \textbf{Minimum simplicity of images for pre-training ViTs.}
We created an extremely simple dataset with images that only contained randomly drawn lines (LineDB). The classes were assigned by counting the number of lines in each image. We list example images of 500 randomly selected classes with 5 instances per class in Figure~\ref{fig:linedb_cat_ins}. The definitions of classes and instances are shown in Table~\ref{tab:fdsl_class_instance}.

In this experiment, we assigned LineDB-\{16, 32, 64, 128, 256, 512, 1,000\} categories in the pre-training phase (see Figure~\ref{fig:linedb}).
LineDB is useful for pre-training. The model pre-trained with LineDB-512 produced accuracies of \{95.6, 77.7, 71.9, 96.8\} on \{C10, C100, Cars, Flowers\} (much better than those from scratch in Table~\ref{tab:comparison}). However, the pre-training effect slightly decreases for 1k categories. This is related to the optimal degree of contour complexity shown in Table~\ref{tab:numof_vertices}.
Moreover, for the dataset to have a positive pre-training effect, consistent labels need to be assigned to non-trivial images.
The performance of LineDB-512 with random permutation was very low, with \{13.5, 1.9, 0.8, 3.4\} on \{C10, C100, Cars, Flowers\}.

% Discussion of Hypotheses 1 and 2
\subsection{Comparisons of SL, SSL, and FDSL}
\label{sec:parameter_tuning_comparison}

In the comparison experiments, we created FDSL methods with the same dataset size as that for the baseline method and compared them after pre-training.

\noindent \textbf{ImageNet-1k fine-tuning (Table~\ref{tab:comparison_imagenet1k}).}
We compared the results for fine-tuning on ImageNet-1k after pre-training on large real and synthetic datasets.
We used RCDB/ExFractalDB with 21k and 50k classes for the pre-training and compared it with ImageNet-21k pre-training.
We fixed the number of instances to 1k and varied the number of classes, as done in previous work~\cite{KataokaACCV2020,NakashimaarXiv2021}, but for a larger number of classes.
For the large datasets, the pre-training was conducted with fewer epochs to keep the total number of images used for pre-training somewhat constant.
For the experiments on larger datasets with more than 21k classes, we used roughly the same hyperparameters and data augmentation as those in the DeiT paper~\cite{TouvronICML2021}, except for the batch size, which we changed from 1,024 to 8,192.
We also varied the learning rate and the number of epochs in response to changes in the size of dataset/model/batch.
Specifically, we set the learning rate for ViT-{Tiny, Base, Large} to {8.0e-3, 1.0e-3, 5.0e-4}, and set the number of epochs for datasets with {21k, 50k, 100k} classes to {90, 40, 20}, respectively.

For the pre-training on larger datasets, we used WebDataset\footnote{WebDataset library: \url{https://github.com/tmbdev/webdataset}} to accelerate input/output processing when loading the dataset.
In addition, to fit large models like ViT-Large into memory, we use Gradient Checkpointing~\cite{chen2016training} to save memory for activation.

We would like to emphasize that we did not perform excessive hyper-parameter tuning or cherry-picking. We used the same hyperparameters from the smaller ExFractalDB when generating the larger datasets, and the pre-training of ViT on these datasets was done with the same hyperparameters as in the DeiT paper. We performed each of the large-scale runs (on \{ImageNet, ExFractal, and RadialContour\}-21k) only once.

The results in Table~\ref{tab:comparison_imagenet1k} show that the ImageNet-1k accuracy for ViT-Base is 79.8 when trained from scratch and 81.8 when pre-trained with ImageNet-21k.
Pre-training with ExFractalDB-21k (82.7), RCDB-21k (82.4), and RCDB-50k (82.6) outperformed that with ImageNet-21k.
The fact that we can match the accuracy of pre-training on ImageNet-21k with synthetic datasets of the same size is rather surprising given that the domain of the synthetic datasets is very different from that of the real images in ImageNet-1k/21k.

\begin{table}[t]
	\begin{center}
	\caption{Comparison of ImageNet-1k fine-tuning. Accuracies obtained with ViT-Tiny/Base architectures are listed. 21k/50k indicates the number of classes in the pre-training phase. Best and second-best values for a given dataset are in underlined bold and bold, respectively.}
    \begin{tabular}{lccccccccc} \toprule[0.8pt]
        Pre-training & Img & Type & ViT-Tiny & ViT-Base\\\midrule[0.5pt]
        From scratch & -- & -- & 72.6 & 79.8 \\
        ImageNet-21k & Real & SL & \underline{\textbf{74.1}} & 81.8 \\
        %JFT-300M & Real & SL & -- & \underline{\textbf{84.1}}$^{*}$ \\
        FractalDB-21k & Synth & FDSL & 73.0 & 81.8 \\
        FractalDB-50k & Synth & FDSL &73.4 & 82.1 \\
        ExFractalDB-21k & Synth & FDSL & 73.6 & \underline{\textbf{82.7}} \\
        ExFractalDB-50k & Synth & FDSL & \textbf{73.7} & 82.5 \\
        RCDB-21k & Synth & FDSL & 73.1 & 82.4 \\
        RCDB-50k & Synth & FDSL & 73.4 & \textbf{82.6} \\
        \bottomrule[0.8pt]
    \end{tabular}
    \label{tab:comparison_imagenet1k}
    \end{center}
\end{table}

% Pre-Train with scaling ExFractalDB and ViT size, Fine-Tune ImageNet-1k.
\begin{table}[t]
	\begin{center}
	\caption{Comparison of ImageNet-1k fine-tuning by models pre-trained with up-scaled ExFractalDB and ViT sizes.}
    \begin{tabular}{lcccc} \toprule[0.8pt]
        Pre-training & Img & Type & Arch. & Acc.\\\midrule[0.5pt]
        ImageNet-21k & Real & SL & ViT-Base & 81.8 \\ \bottomrule[0.5pt]
        ExFractalDB-21k & Synth & FDSL & ViT-Base & 82.7 \\
        ExFractalDB-50k & Synth & FDSL & ViT-Base & 82.5 \\
        ExFractalDB-100k & Synth & FDSL & ViT-Base & 82.7 \\ \bottomrule[0.5pt]
        ExFractalDB-21k & Synth & FDSL & ViT-Large & 80.4 \\
        ExFractalDB-50k & Synth & FDSL & ViT-Large & 81.0 \\
        \bottomrule[0.8pt]
    \end{tabular}
    \label{tab:comparison_imagenet1k_with_scaling_dataset_and_model}
    \end{center}
\end{table}

% Fine-Tune ImageNet-1k with High resolution.
\begin{table}[t]
    \begin{center}
	\caption{Comparison of ImageNet-1k fine-tuning with different resolutions. Accuracies were obtained with the ViT-Base architecture.}
    \begin{tabular}{lcccc} \toprule[0.8pt]
        Pre-training & Img & Type & FT Resolution & Acc. \\ \midrule[0.5pt]
        ImageNet-21k & Real & SL & 224$^{2}$ & 81.8 \\
        ImageNet-21k & Real & SL & 384$^{2}$ & 83.0 \\
        \rowcolor[gray]{0.8} JFT-300M\cite{DosovitskiyICLR2021} & Real & SL & 384$^{2}$ & \underline{\textbf{84.1}}$^{*}$ \\
        ExFractalDB-21k & Synth & FDSL & 224$^{2}$ & 82.7 \\
        \rowcolor[gray]{0.8} ExFractalDB-21k & Synth & FDSL & 384$^{2}$ & \textbf{83.8} \\
        % ExFractalDB-21k & Synth & FDSL & 512$^{2}$ & 83.9 \\
        RCDB-21k & Synth & FDSL & 224$^{2}$ & 82.4 \\
        \rowcolor[gray]{0.8} RCDB-21k & Synth & FDSL & 384$^{2}$ & 83.5 \\
        VisualAtom-21k & Synth & FDSL & 224$^{2}$ & 82.7 \\
        \rowcolor[gray]{0.8} VisualAtom-21k & Synth & FDSL & 384$^{2}$ & 83.7 \\
        \bottomrule[0.8pt]
        \multicolumn{5}{l}{* Rate reported in original ViT paper~\cite{DosovitskiyICLR2021}.}\\
        %\bottomrule[0.8pt]
    \end{tabular}
    \label{tab:comparison_imagenet1k_with_high_resolution}
	\end{center}
\end{table}

\begin{table*}[t]
	\begin{center}
	\caption{Comparison of object detection and instance segmentation. Several pre-trained models were validated on the COCO dataset. We list and compare the models pre-trained for 36 and 60 epochs. Best values for each learning type are in bold.}
    \begin{tabular}{lcccc} \toprule[0.8pt]
        Pre-training & Type & \#FT-epoch &  COCO Det & COCO Inst Seg \\
         &  &  & AP$_{50}$ / AP / AP$_{75}$ & AP$_{50}$ / AP / AP$_{75}$ \\
        \bottomrule[0.5pt]
        From scratch & -- & 36 & 64.6 / 43.4 / 47.6 & 61.4 / 39.1 / 42.0 \\
        ImageNet-1k & SL & 36 & 70.2 / 48.6 / 53.0 & 67.5 / 43.4 / 46.4 \\
        ImageNet-21k & SL & 36 & \textbf{71.2} / \textbf{49.0} / \textbf{53.9} & \textbf{68.4} / \textbf{44.0} / \textbf{47.5} \\
        \bottomrule[0.5pt]
        ExFractalDB-1k & FDSL & 36 & 68.2 / 47.0 / 51.6 & 65.4 / \textbf{42.0} / \textbf{45.3} \\
        ExFractalDB-21k & FDSL & 36 & \textbf{68.7} / \textbf{47.3} / \textbf{51.9} & \textbf{65.6} / \textbf{42.0} / 45.0 \\
        RCDB-1k & FDSL & 36 & 67.1 / 46.0 / 50.6 & 64.2 / 41.0 / 44.0 \\
        RCDB-21k & FDSL & 36 & 65.7 / 44.4 / 48.7 & 62.5 / 39.7 / 42.7 \\
        \bottomrule[0.8pt]
        From scratch & -- & 60 & 63.7 / 42.2 / 46.1 & 60.7 / 38.5 / 41.3 \\
        ImageNet-1k & SL & 60 & 69.2 / 48.2 / 53.0 & 66.6 / 43.1 / 46.5 \\
        ImageNet-21k & SL & 60 & \textbf{70.7} / \textbf{48.8} / \textbf{53.2} & \textbf{67.7} / \textbf{43.6} / \textbf{47.0} \\
        \bottomrule[0.5pt]
        ExFractalDB-1k & FDSL & 60 & 69.1 / \textbf{48.0} / \textbf{52.8} & 66.3 / \textbf{42.8} / 45.9 \\
        ExFractalDB-21k & FDSL & 60 & \textbf{69.2} / \textbf{48.0} / 52.6 & \textbf{66.4} / \textbf{42.8} / \textbf{46.1} \\
        RCDB-1k & FDSL & 60 & 68.3 / 47.4 / 51.9 & 65.7 / 42.2 / 45.5 \\
        RCDB-21k & FDSL & 60 & 67.7 / 46.6 / 51.2 & 64.8 / 41.6 / 44.7 \\
        \bottomrule[0.8pt]
    \end{tabular}
    \label{tab:comparison_detection_segmentation}
    \end{center}
\end{table*}

\begin{table*}[t]
    \begin{center}
    \caption{Comparison of pre-training for SL/SSL methods. For SSL, (D) indicates  DINO~\cite{CaronICCV2021_dino}. Best values for each learning type are in bold.}
    \begin{tabular}{lccccccccc|c} \toprule[0.8pt]
        Pre-training & Img & Type & C10 & C100 & Cars & Flowers & VOC12 & P30 & IN100 & Average \\\midrule[0.5pt]
        From scratch & -- & -- &  78.3 & 57.7 & 11.6 & 77.1 & 64.8 & 75.7 & 73.2 & 62.6 \\
        Places-365 & Real & SL & 97.6 & 83.9 & 89.2 & 99.3 & 84.6 & -- & 89.4 & -- \\
        ImageNet-1k & Real & SL & \textbf{98.0} & \textbf{85.5} & \textbf{89.9} & \textbf{99.4} & \textbf{88.7} & \textbf{80.0} & -- & -- \\\midrule[0.5pt]
        ImageNet-1k & Real & SSL (D) & \textbf{97.7} & 82.4 & \textbf{88.0} & 98.5 & 74.7 & 78.4 & \textbf{89.0} & 86.9 \\
        PASS & Real & SSL (D) & 97.5 & \textbf{84.0} & 86.4 & \textbf{98.6} & \textbf{82.9} & \textbf{79.0} & 82.9 & \textbf{87.8} \\\midrule[0.5pt]
        FractalDB-1k~\cite{NakashimaarXiv2021} & Synth & FDSL & 96.8 & 81.6 & 86.0 & 98.3 & 80.6 & 78.4 & 88.3 & 87.1 \\
        %FractalDB-10k & Synth & FDSL & \textbf{97.8} & 83.1 & 89.1 & 98.8 & 82.6 & \textbf{80.8} & 88.0 \\
        RCDB-1k & Synth & FDSL & 97.0 & 82.2 & 86.5 & 98.9 & 80.9 & \textbf{79.7} & 88.5 & 87.6 \\
        %RCDB-10k & Synth & FDSL & 97.0 & 83.5 & 87.4 & 99.2 & 81.6 & \underline{\textbf{81.5}} & 89.8 \\
        ExFractalDB-1k & Synth & FDSL & 97.2 & 81.8 & 87.0 & \textbf{98.9} & 80.6 & 78.0 & 88.1 & 87.4 \\
        %ExFractalDB-10k & Synth & FDSL & 97.6 & \textbf{84.3} & 89.2 & \textbf{99.4} & 82.4 & 81.3 & \underline{\textbf{89.9}} \\
        \rowcolor[gray]{0.8} ExFractalDB-1k* & Synth & FDSL & \textbf{97.5} & \textbf{82.6} & \textbf{90.3} & \textbf{99.6} & \textbf{81.4} & 79.4 & \textbf{89.2} & \underline{\textbf{88.6}} \\
        \bottomrule[0.8pt]
        \multicolumn{11}{l}{* Rate calculated for 1.4M images, which is the same number of images in the PASS dataset.}\\
    \end{tabular}
    \label{tab:comparison}
    \end{center}
\end{table*}

\begin{table}[t]
    \begin{center}
    \caption{Comparison of ViT, gMLP, and ResNet with FractalDB-1k, ExFractalDB-1k, RCDB-1k, and LineDB-1k pre-training.} %Best values are in bold.
    \begin{tabular}{lccccc} \toprule[0.8pt]
        Pre-training & Arch. & C10 & C100 & Cars & Flowers \\ \midrule[0.5pt]
        FractalDB & ResNet & 95.7 & 79.0 & 80.9 & 96.9 \\ 
        FractalDB & gMLP & 95.4 & 77.4 & 78.7 & 94.2 \\
        \rowcolor[gray]{0.8} FractalDB & ViT & \textbf{96.8} & \textbf{81.6} & \textbf{86.0} & \textbf{98.3} \\
        ExFractalDB & ResNet & 96.1 & 80.4 & 80.3 & 97.4 \\ 
        ExFractalDB & gMLP & 96.7 & 80.0 & 84.5 & 98.6 \\ 
        \rowcolor[gray]{0.8} ExFractalDB & ViT & \textbf{97.2} & \textbf{81.8} & \textbf{87.0} & \textbf{98.9} \\
        RCDB & ResNet & 95.6 & 78.4 & 71.6 & 94.2 \\ 
        RCDB & gMLP & 96.1 & 78.6 & 78.1 & 96.6 \\ 
        \rowcolor[gray]{0.8} RCDB & ViT & \textbf{97.0} & \textbf{82.2} & \textbf{86.5} & \textbf{98.9} \\
        LineDB & ResNet & 91.8 & 65.5 & 15.6 & 71.1 \\ 
        LineDB & gMLP & 93.9 & 73.2 & 30.2 & 85.3 \\ 
        \rowcolor[gray]{0.8} LineDB & ViT & \textbf{95.6} & \textbf{77.7} & \textbf{71.9} & \textbf{96.8} \\
        \bottomrule[0.8pt]
        %DeiT-Ti/16 & 5 & 96.8 & 81.6 & 86.0 & 98.3 \\\bottomrule[0.8pt]
        % DeiT-B/16 & 86 & 97.1 & 83.2 & 86.5 & 97.9 \\ 
    \end{tabular}
    \label{tab:cnn_mlp_vit}
    \end{center}
\end{table}

\noindent \textbf{Scaling ExFractalDB size and ViT size (Table~\ref{tab:comparison_imagenet1k_with_scaling_dataset_and_model}).}
We further evaluated the ExFractalDB-21k pre-training that recorded the highest fine-tuning accuracy in Table~\ref{tab:comparison_imagenet1k}, in terms of both pre-training dataset and model scaling.
Table~\ref{tab:comparison_imagenet1k_with_scaling_dataset_and_model} shows that scaling the pre-training dataset without changing the settings including model size and learning rate, does not improve accuracy (from ExFractalDB-21k to ExFractalDB-50k/100k in ViT-Base). 
This is also known for pre-training with real image datasets~\cite{DosovitskiyICLR2021}.
The accuracy of fine-tuning is improved when ExFractalDB is scaled from 21k to 50k with not ViT-Base but ViT-Large.
The smaller accuracy of ViT-Large compared to ViT-Base is likely due to a need to adjust the hyperparameters such as learning rate and batch size.

\noindent \textbf{ImageNet-1k fine-tuning with high resolution input (Table~\ref{tab:comparison_imagenet1k_with_high_resolution}).}
Table~\ref{tab:comparison_imagenet1k_with_high_resolution} shows that the comparison of ImageNet-1k fine-tuned with different resolutions. 
The experimental setting followed that in the original ViT paper~\cite{DosovitskiyICLR2021}. Here, we fairly compare the pre-training on ImageNet-21k and JFT-300M by the method of Dosovitskiy \textit{et al.}
In this result, with a resolution of $384^{2}$ pixels at fine-tuning, the model pre-trained on ExFractalDB-21k (83.8) is a close match to the reported performance of JFT-300M (84.1)~\cite{DosovitskiyICLR2021}. Similarly, pre-training on RCDB-21k (83.5) also improves accuracy by up-scaling resolution during fine-tuning.
Increasing the resolution of images input to the model during fine-tuning enhanced the accuracy.

\noindent \textbf{COCO detection/instance segmentation (Table~\ref{tab:comparison_detection_segmentation}).}
We additionally validated FDSL on detection and instance segmentation tasks using the COCO dataset~\cite{LinECCV2014_coco}. We used a Swin Transformer Base~\cite{LiuICCV2021_Swin} backbone, a Mask R-CNN~\cite{HeICCV2017_mask} head, and fine-tuning for 36 and 60 epochs. Our pre-trained model achieved higher scores at all settings in Table~\ref{tab:comparison_detection_segmentation}. The models fine-tuned for 60 epochs yielded similar average precision (AP) to the ones fine-tuned for 36 epochs. This suggests that the FDSL pre-training takes longer than the SL and real-image pre-training. Moreover, although our best pre-trained model (ExFractalDB-21k) did not reach the AP with the ImageNet-21k pre-trained model, the AP was similar to that of the ImageNet-1k pre-trained model. It is not enough to adapt the transfer to object detection and instance segmentation with only hypotheses 1 and 2. We should prepare a more suitable pre-training task for object detection and instance segmentation.

\noindent \textbf{FDSL versus SSL/SL (Table~\ref{tab:comparison}).}
For the following comparisons, we extended the datasets used for fine-tuning.
In addition to C10/100, Cars, and Flowers, we included ImageNet-100 (IN100)~\cite{KataokaACCV2020}, Places30 (P30)~\cite{KataokaACCV2020}, and Pascal VOC 2012 (VOC12)~\cite{EveringhamIJCV2015_voc}. We compared RCDB and FractalDB with DINO on ImageNet-1k and PASS~\cite{AsanoNeurIPS2021_pass}, and human annotations on Places-365 and ImageNet-1k in Table~\ref{tab:comparison}.

A comparison of the average accuracy across all fine-tuning datasets indicates that ExFractalDB-1k with 1.4k instances achieved a higher average accuracy (88.6) than that of the self-supervised PASS (PASS+DINO 87.8).
Both PASS and FDSL {are attempts} to improve ethics in datasets. FDSL shows that it is possible to achieve higher accuracy with synthetic datasets of the same size.

FDSL pre-training partially outperformed ImageNet-1k pre-training in Cars with ExFractalDB-1k (90.3 vs. 89.9) and Flowers with ExFractalDB (99.6 vs. 99.4). Although FDSL did not outperform ImageNet pre-training for all cases, it is competitive across a wide range of fine-tuning.

\noindent \textbf{Performance on gMLP and ResNet.}
Table~\ref{tab:cnn_mlp_vit} shows the results for gMLP~\cite{LiuarXiv2021} and ResNet~\cite{HeCVPR2016}. We used gMLP-Tiny with a 16$\times$16 patch and ResNet with 50 layers. These architectures contain 6.0M and 25.0M parameters, respectively (ViT-Tiny has 5.0M). According to the results, ViT seems to be a better match with FDSL than gMLP and ResNet.

\begin{figure*}[h]
  \centering
  \includegraphics[width=0.9\linewidth]{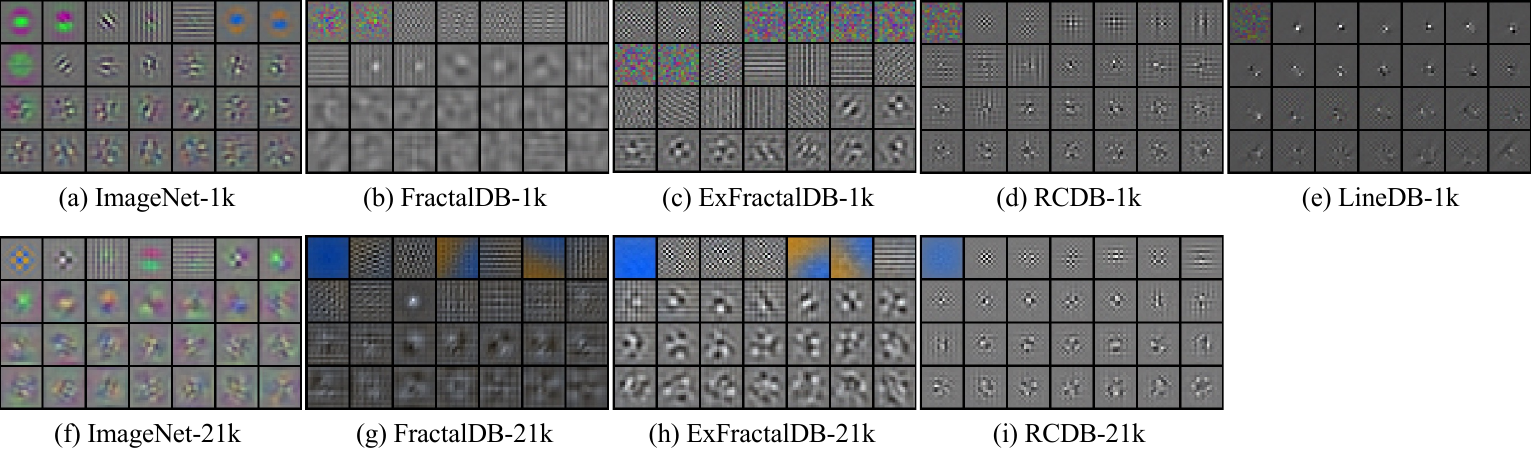}
  \caption{Embedding filters in ImageNet-1k/21k, FractalDB-1k/21k, ExFractalDB-1k/21k, RCDB-1k/21k, and LineDB-1k.}
  \label{fig:embedding_filters}
\end{figure*}

\noindent \textbf{Visualization of attention maps.}
We show the embedding filters for ImageNet-1k/21k, FractalDB-1k/21k, ExFractalDB-1k/21k, RCDB-1k/21k, and LineDB-1k in Figure~\ref{fig:embedding_filters}.

\section{Discussion and Conclusion}
We investigated the possibility of using various forms of FDSL for pre-training ViTs.
We extended the original FractalDB to formulas that focus purely on object contours and increased the complexity of the formula supervision to observe the effect of these changes.
One of our major findings is that we can surpass the accuracy of a ViT pre-trained on ImageNet-21k and be closer fine-tuning accuracy with JFT-300M with only $\times$14.2 fewer images, using our ExFractralDB-21k and RCDB-21k datasets.

In the present work, we provided empirical evidence that supports our two hypotheses. We created a variety of synthetic datasets with different characteristics. When these datasets were used for the pre-training of a ViT, the ones that consist primarily of contours gave the highest fine-tuning accuracy, which validates our first hypothesis. We also controlled the difficulty of the pre-training by varying the number of FDSL parameters. We found that more difficult pre-training tasks led to better fine-tuning accuracy, which validates our second hypothesis.

\noindent \textbf{Discussion of hypotheses 1 and 2.} Here, we summarize the results of hypotheses 1 and 2.

\textit{Hypothesis 1: Object contours are what matter in FDSL datasets.}
The preliminary visualization experiments shown in Figures \ref{fig:figure1} and \ref{fig:contour_selfattention} show that the self-attention is focused on the contours during ViT pre-training. This led to our first hypothesis that object contours are what matter in the dataset for pre-training ViTs. To validate this hypothesis, we constructed various synthetic datasets. From Table \ref{tab:numof_vertices}, we saw that datasets that consist mostly of contour lines, such as BezierCurveDB, RCDB, and FractalDB, had the highest accuracy. This supports our hypothesis that object contours are indeed essential for pre-training ViTs.

Figure \ref{fig:result_fractal_contour} shows the effect of varying the number of points to render FractalDB. We see that using fewer than 50k points results in broken contour lines, for which the pre-training fails. We also broke the contour lines in RCDB by drawing white lines over the shapes, as shown in Figure \ref{fig:rcdb_image_corruption}. From Figure \ref{fig:corruption_rcdb}, we see that this also prevents the ViT from learning a good visual representation.

Table~\ref{tab:multiple_parameters} shows that increasing the number of vertices in RCDB beyond 203--302 makes the pre-training fail completely. In contrast, using fewer vertices while augmenting the number of classes by introducing extra parameters resulted in a significant improvement in the accuracy, even up to 50k classes, as shown in Table~\ref{tab:comparison_detection_segmentation}.
Figure \ref{fig:linedb} shows that the accuracy of LineDB also decreases when the number of lines exceeds 512.

\textit{Hypothesis 2: Increasing the number of parameters in FDSL pre-training.}  
Table~\ref{tab:fdb1k_labels} shows that the pre-training with FractalDB results in higher scores than pre-training using external labels with SSL.
In addition, it was found that the more fractal parameters were varied (three and six parameters are compared in Table 1), the better the pre-training effect was.

For FractalDB, we extended the IFS from 2D to 3D and increased the number of parameters in the equation from 6 to 12, which led to a significant increase in the number of classes, each with distinct features.
When projecting a 3D fractal onto a 2D image, we used a random perspective instead of a set of fixed perspectives.

For RCDB, we varied the parameter set (Table~\ref{tab:rcdb}), including the number of contours, radius, line width, resizing factor, and Perlin noise, in addition to the number of vertices, with each combination categorized as a different class. The parameters were determined by the exploration in Figure~\ref{fig:rcdb_exploration}.
For RCDB, we varied the parameter set $\eta$ with each combination categorized as a different class.

Table~\ref{tab:multiple_parameters} shows that each of these modifications led to a notable improvement in accuracy.
For RCDB, Table~\ref{tab:comparison_imagenet1k} shows that the increase in accuracy with an increasing number of classes continues up to 50k classes.

\noindent \textbf{Potential of FDSL.} We believe that SL on synthetic images (FDSL) is a broad area of research at a similar granularity as that of SSL. In the same way that numerous papers on SSL were published after the initial work, we believe that this new mode of training (FDSL) deserves the same kind of attention and allocation of slots for the community to extract its full potential. Moreover, there have been some concerns that such simplistic synthetic images may work for small dataset sizes shown in the existing work, but will not be able to scale up to the scale of ImageNet-21k pre-training of ViTs. Our current results show for the first time that it does indeed work at this scale, which is an important finding.

However, as shown in Tables~\ref{tab:comparison_imagenet1k} and \ref{tab:comparison_imagenet1k_with_scaling_dataset_and_model}, dataset scaling does not necessarily increase accuracy in FDSL pre-training. Especially in the ExFractalDB pre-training, the fine-tuning accuracy seems to saturate at ExFractalDB-21k. The resulting accuracies are 82.5 (ExFractalDB-50k) and 82.7 (ExFractalDB-100k). To this point, Table~\ref{tab:comparison_imagenet1k_with_high_resolution} shows a clue to a breakthrough to increase the pre-training effect in FDSL datasets. The ImageNet-1k fine-tuning was done by following the original ViT paper, namely the image resolution was changed from $224^2$ pixels to $384^2$ pixels. Surprisingly, the fine-tuning accuracy with ExFractalDB-21k (83.8) is similar to the accuracy with a JFT-300M pre-trained model (84.1). Note that our proposed pre-training is done by a $\times14.2$ relatively smaller dataset in comparison to JFT-300M  (ExFractalDB-21k: 21M images vs. JFT-300M: 300M images). We believe that further improvements in contour shapes and a more complex classification task are possible, which leaves open the possibility to scale up the pre-training on synthetic datasets to one day outperform JFT-300M/3B~\cite{SunICCV2017_jft300m,ZhaiarXiv2021_ScalingViT} and IG-3.5B~\cite{MahajanECCV2018_ig3.5b}.

Relatively small datasets on the order of 1k categories have been enhanced with hypotheses 1 and 2. Even using ExFractalDB-1k pre-training, we can match the training of well-known real-image datasets and SSL labels. In Table~\ref{tab:comparison}, the average accuracy in ExFractalDB-1k with 1.4k instances per category recorded a better than the average accuracy in the self-supervised PASS dataset (ExFractalDB-1k: 88.6 vs. PASS: 87.8). Moreover, the ExFractalDB-1k pre-trained model partially surpassed the ImageNet-1k pre-trained model in fine-tuning datasets on Cars and Flowers.

\noindent \textbf{Better pre-training for AI models.} We have been seeking better neural network pre-training for image recognition. Not limited to image classification as mentioned in the present paper, we also performed object detection and instance segmentation trials. However, the as-is pre-training method by image classification did not reach the performance level of ImageNet-1k/21k with real images and manual annotations. We consider that a more similar pre-training method should be implemented to improve the localization tasks in object detection and instance segmentation. To this point, formula-driven approaches in video recognition~\cite{KataokaWACV2022_vpn}, multi-view image recognition~\cite{YamadaIROS2021}, and 3D point cloud detection~\cite{YamadaCVPR2022_pcfractal} have introduced an effective pre-training method at each modality and task. Moreover, in natural language processing, an artificial language generated from Zipf's law makes an effective task for a language model~\cite{RiACL2022_artificiallanguage}. To enhance the fine-tuning performance, it is better that a pre-training task is adjusted along with the fine-tuning task. No matter what the data modality, we can employ the FDSL approach to construct an AI model. The FDSL approach allows us to automatically generate labeled data samples from a principle. We consider that any pre-training methods can be replaced by synthetic data.

\ifCLASSOPTIONcompsoc
  \section*{Acknowledgments}
    This paper is based on results obtained from a project, JPNP20006, commissioned by the New Energy and Industrial Technology Development Organization (NEDO). AIST policy-based budget project ``R\&D on Generative AI Foundation Models for the Physical Domain.'' Computational resource of AI Bridging Cloud Infrastructure (ABCI) provided by National Institute of Advanced Industrial Science and Technology (AIST) was used. We want to thank Junichi Tsujii, Yutaka Satoh, Kensho Hara, Yoshihiro Fukuhara, Hiroaki Aizawa, Shintaro Yamamoto, Takehiko Ohkawa, Ryo Takahashi for their helpful comments in research discussions.
\else
  \section*{Acknowledgment}
\fi

\ifCLASSOPTIONcaptionsoff
  \newpage
\fi

\bibliographystyle{IEEEtran}
\bibliography{egbib}

\end{document}